%% file: main.tex
\pdfoutput=1

\documentclass[10pt,twocolumn,letterpaper]{article}

\usepackage{cvpr}              %

\input{preamble}

\usepackage{cuted}
\usepackage{caption}
\usepackage{multirow}
\usepackage{float}
\usepackage{colortbl}

\definecolor{cvprblue}{rgb}{0.21,0.49,0.74}
\usepackage[pagebackref,breaklinks,colorlinks,allcolors=cvprblue]{hyperref}

\def\confName{CVPR}
\def\confYear{2027}

\title{PINE: Point, Instance and Normal Estimation in One-Step Diffusion Model}

\author{
\raisebox{6pt}[0pt][0pt]{\begin{minipage}[t]{\textwidth}\centering
Zeyu Ren$^{1}$ \quad
Jie Ni$^{2*}$ \quad
Rui Qian$^{3*}$ \quad
Jiahao Wei$^{3}$ \quad
Hao Tang$^{4}$ \quad
Yongtai Liu$^{5}$ \quad
Zeyu Zhang$^{4\dag}$ \\
$^{1}$The University of Melbourne \quad
$^{2}$Southeast University \quad
$^{3}$Fudan University \\
$^{4}$Peking University \quad
$^{5}$Hanyang University \\
\small $^*$Corresponding authors: njie@seu.edu.cn, qiianruii@gmail.com.
$^\dag$Project lead.
\end{minipage}}%
}

\begin{document}
\maketitle

\begin{strip}
\vspace{-2.4em}
    \centering
    \addtolength{\abovecaptionskip}{3pt}%
    \addtolength{\belowcaptionskip}{-3pt}%
    \raisebox{-10pt}[\height][\depth]{\includegraphics[width=\textwidth]{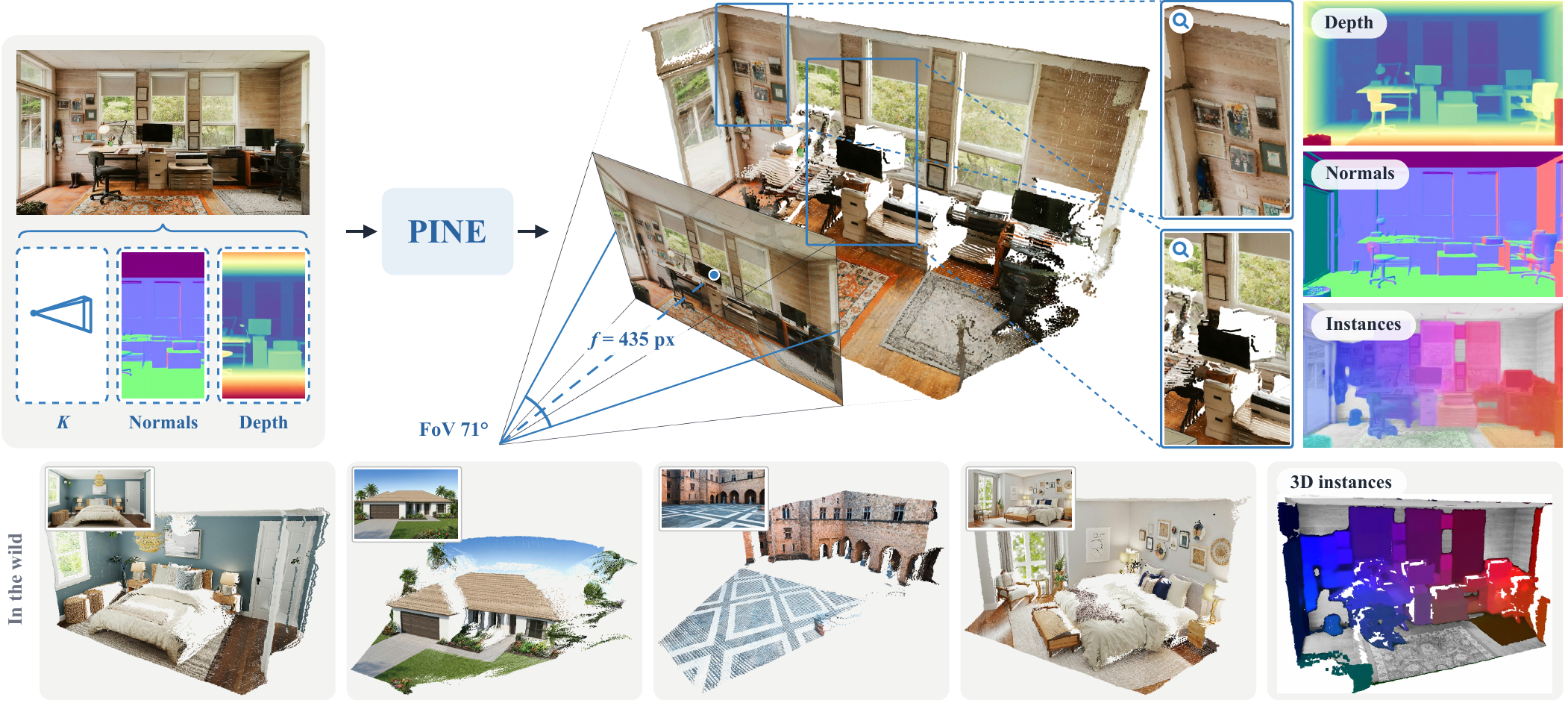}}
    \captionof{figure}{\textbf{PINE} predicts the point map, the normals and the instance map of an image in one forward pass.
    With limited supervision, it achieves state-of-the-art point maps and runs 5.7$\times$ faster than MoGe-3 combined with SAM.}
    \label{fig:teaser}
\end{strip}

\begin{abstract}
Understanding a scene requires both geometry and semantics: geometry recovers its spatial structure, and semantics separates it into objects.
Current models obtain the two from separate networks or separate passes.
We present \textbf{PINE}, a one-step Omni Diffusion Transformer that predicts the point map, the surface normals, and the instance map of an image in one forward pass.
PINE treats each map as a token stream, and a stream mask decides which streams interact.
A camera-aware ray encoding positions every token by its camera ray rather than its pixel.
Grounding attention adds the affinity of pretrained DINOv3 features to the attention logits.
Trained on only 139K synthetic images for geometry, PINE achieves state-of-the-art point maps whether or not intrinsics are given and the best average rank on surface normals.
It also exceeds SAM on large objects with less than \textbf{1\%} of its training images, and it predicts all three maps \textbf{5.7$\times$ faster} than MoGe-3 with SAM.
Website: \url{https://aigeeksgroup.github.io/PINE}
\end{abstract}

\section{Introduction}
\label{sec:intro}

Understanding a scene requires both geometry and semantics~\cite{putting,taskonomy}.
Geometry describes where each surface lies and how it is oriented, and semantics describes which surfaces form one object.
Specialist models~\cite{depth_survey, instance_survey} now recover each part from a single image with high accuracy.
The Depth Anything~\cite{da,da2,da3} and MoGe~\cite{moge,moge2,moge3} series predict depth, point map and normals.
Diffusion estimators such as Marigold~\cite{marigold,marigoldv2} and Lotus~\cite{lotus} predict detailed depth and surface normals, and GenPercept~\cite{genpercept} and E2E-FT~\cite{e2eft} reduce them to one sampling step.
The SAM series~\cite{sam,sam2} segments any object under a prompt, and gen2seg~\cite{gen2seg} colors every instance with a generative model.

None of these models predicts all three maps in one forward pass.
A full parse of one image chains a geometry model and a segmentation model.
Their latencies and memory add up, and their maps do not share one camera.
Point-map models~\cite{moge2,moge3,foundationgeo,unidepthv2,da3} can take the intrinsics, whereas diffusion normal estimators~\cite{marigold,geowizard,stablenormal,lotus,e2eft} take the image alone, even when the intrinsics are given.
Generalist models such as 4M~\cite{4m}, Unified-IO~\cite{unifiedio}, DICEPTION~\cite{diception} and OneDiffusion~\cite{onediffusion} cast multiple tasks into one network.
They do not remove the cost: DICEPTION runs one forward pass per map, and OneDiffusion needs multiple sampling steps for each map.
A direct solution is one Diffusion Transformer that predicts all maps together in one step.
Yet a Diffusion Transformer~\cite{dit} is designed for RGB images.
The three maps break three of its assumptions.
The instance map has no fixed target, because it is correct under any permutation of its colors.
The point map places each depth along the camera ray of its pixel, but a transformer indexes its tokens by pixel~\cite{dit}.
A full view and a zoomed crop then give one scene point two positions.
The normals and the instance map are constant over surfaces and objects, but attention recovers this grouping from the tokens alone.
In practice, the attention of a patch spreads over other objects.

Based on these observations, we present PINE, a one-step Omni Diffusion Transformer that predicts the point map, the normals and the instance map in one forward pass, as \cref{fig:teaser} shows.
PINE revises these three assumptions.
For the target, PINE represents each map as a token stream.
The geometry streams follow flow matching at their own timesteps, while the instance stream regresses its map with a clustering loss.
A stream mask keeps the instance stream apart from the geometry noise.
For the position, the camera-aware ray encoding positions the tokens of every stream by the rays of one camera.
When the intrinsics are unknown, a default camera takes their place.
For the grouping, grounding attention writes the affinity of DINOv3 features~\cite{dinov3} into the attention logits of the normal and instance streams.
In summary, our contributions are as follows:
\begin{itemize}
\item We present PINE, a one-step Omni Diffusion Transformer. One forward pass replaces one specialist model per map and multiple sampling steps of diffusion estimators.
\item We introduce a camera-aware ray encoding, which positions tokens by camera rays whether or not intrinsics are given, and grounding attention, which gives the normal and instance streams the grouping of their pixels.
\item PINE achieves state-of-the-art point maps whether or not intrinsics are given and the best average rank on surface normals, and it exceeds SAM on large objects with less than 1\% of its training images. It runs 5.7 times faster than MoGe-3 combined with SAM.
\end{itemize}

\section{Related Work}
\label{sec:related}

\begin{figure*}[t]
    \centering
    \includegraphics[width=\textwidth]{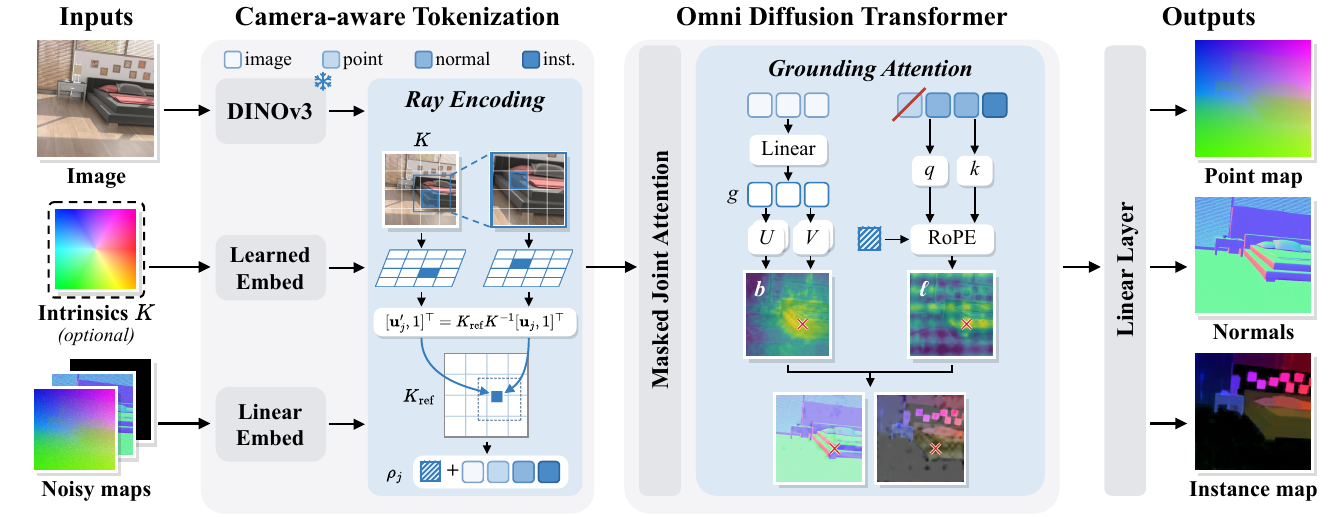}
    \caption{\textbf{Overview of PINE.}
    The point, normal and instance streams embed their maps with the frozen DINOv3 features and pass through DiT blocks as one sequence.
    In attention, the stream mask decides which streams interact, ray positions encode the intrinsics and grounding attention adds the DINOv3 affinity.
    Each stream decodes its own map, and one step at inference gives all three maps.}
    \label{fig:overview}
\end{figure*}

\paragraph{Monocular Geometry Estimation.}
MiDaS~\cite{midas,dpt} and Depth Anything~\cite{da,da2} train on mixed or large unlabeled data to predict affine-invariant depth.
DUSt3R~\cite{dust3r} regresses point maps directly, and MoGe~\cite{moge,moge2,moge3} predicts affine-invariant point maps from a single image.
Without intrinsics, the field of view is ambiguous.
Metric3D~\cite{metric3d}, UniDepth~\cite{unidepth,unidepthv2}, Depth Pro~\cite{depthpro}, DA3~\cite{da3} and FoundationGeo~\cite{foundationgeo} estimate the intrinsics or take them as input.
DSINE~\cite{dsine} feeds per-pixel ray directions to a normal estimator.
All of these models index their features by pixels and predict geometry only.
GTA~\cite{gta} and PRoPE~\cite{prope} encode the relative cameras of multiple views in attention, whereas PINE positions the tokens of one image by their rays.

\paragraph{Diffusion for Dense Prediction.}
Marigold~\cite{marigold} fine-tunes a pretrained latent diffusion model~\cite{ddpm,ldm} into a zero-shot depth estimator.
GeoWizard~\cite{geowizard}, StableNormal~\cite{stablenormal} and Marigold V2~\cite{marigoldv2} extend this recipe to normals.
FE2E~\cite{fe2e} starts from an image editing model.
GeoNeXt~\cite{geonext} starts from a video generative model and predicts depth and normals as the next frames.
GenPercept~\cite{genpercept}, Lotus~\cite{lotus} and E2E-FT~\cite{e2eft} reduce the multiple sampling steps of these estimators to one deterministic step.
TransNormal-2~\cite{transnormal2} corrects the decoding errors of the autoencoder at object boundaries with pixel-space losses and a refinement module.
PPD~\cite{ppd}, PointDiT~\cite{pointdit} and Lapis~\cite{lapis} avoid the detail loss of the autoencoder by denoising in pixel space~\cite{jit}.
Each of these estimators predicts one or two geometric maps, whereas PINE predicts all three maps in one step.

\paragraph{Instance Segmentation.}
Instance segmentation has moved from closed label sets~\cite{maskrcnn,mask2former} to class-agnostic entities~\cite{entityseg}.
SimpleClick~\cite{simpleclick} segments the object under a click.
The SAM series~\cite{sam,sam2} accepts any prompt and scales this ability with a large mask corpus.
TokenCut~\cite{tokencut}, Deep Spectral Methods~\cite{DeepSpectralMethods} and CutLER~\cite{cutler} group pixels into objects by the affinities of self-supervised features~\cite{dinov2,dinov3}.
gen2seg~\cite{gen2seg} fine-tunes a generative model~\cite{ldm} to color every instance with a clustering loss~\cite{instanceloss}.
PINE predicts instances as one stream of a geometry model and writes the affinity of DINOv3 features into its attention.

\paragraph{Omni Models for Dense Prediction.}
Multi-task networks such as UberNet~\cite{ubernet} and Omnidata~\cite{omnidata} share one backbone across geometric and semantic tasks.
Taskonomy~\cite{taskonomy} measures how these tasks transfer.
Generalist models cast multiple tasks into one interface, such as images in Painter~\cite{painter} and token sequences in 4M~\cite{4m} and Unified-IO~\cite{unifiedio}.
DICEPTION~\cite{diception} trains one diffusion model for all tasks but runs one forward pass per task.
GenCeption~\cite{genception} turns a video generation model into one perception model, where a text prompt selects the task.
UniDiffuser~\cite{unidiffuser} and OneDiffusion~\cite{onediffusion} give each modality or view its own timestep, so that any of them can condition the others over multiple sampling steps.
PINE is an omni model of a different form: each map is a token stream, each geometry stream has its own timestep, and one forward pass gives all maps.

\section{Method}
\label{sec:method}

We introduce PINE, a one-step diffusion model that takes an RGB image and, optionally, its intrinsics, and predicts an affine-invariant point map, surface normals and an instance map.
\Cref{fig:overview} shows its pipeline.
The Omni Diffusion Transformer of \cref{sec:joint_dit} processes the three maps as token streams in one sequence.
The camera-aware ray encoding of \cref{sec:ray_encoding} positions every token by its camera ray.
Grounding attention in \cref{sec:grounding} adds the DINOv3 affinity to the attention logits of the normal and instance streams.
\Cref{sec:objective} describes the training, and \cref{alg:pine} lists one training step and the one-step inference.

\subsection{Omni Diffusion Transformer}
\label{sec:joint_dit}

Given an RGB image $I$ and, optionally, its intrinsic matrix $K$, PINE predicts the three maps as
\begin{equation}
(P, N, C) = \Phi_\theta(I, K),
\label{eq:task}
\end{equation}
where $P$ is an affine-invariant point map whose $z$ channel is the depth, $N$ holds the surface normals and $C$ is an instance map that assigns every object its own color.
$P$ and $N$ have one correct value per pixel, whereas $C$ is correct under any permutation of its colors and has no fixed target to denoise toward.
We therefore learn $P$ and $N$ by flow matching and $C$ with a clustering loss, and $\Phi_\theta$ represents the three maps as token streams to combine both objectives.

\paragraph{Pixel-Space Diffusion.}
Latent diffusion estimators~\cite{marigold,geowizard} decode their maps through an autoencoder, whose lossy reconstruction smooths depth edges into flying points~\cite{ppd,pointdit}.
PINE denoises the maps in pixel space and avoids this loss.
We index the streams by $s \in \{\text{p}, \text{n}, \text{i}\}$.
Each geometry map $x_s$, with $x_\text{p} = P$ and $x_\text{n} = N$, is noised at timestep $t_s \in [0, 1]$ as
\begin{equation}
z_s = t_s\, x_s + (1 - t_s)\, \sigma \epsilon_s, \qquad \epsilon_s \sim \mathcal{N}(0, \mathbf{I}),
\label{eq:interp}
\end{equation}
where $t_s = 1$ gives the clean map and $\sigma$ is the noise scale.
The network predicts the clean maps $\hat{x}_s$ from the inputs of all streams, and the predicted and target velocities are
\begin{equation}
\hat{v}_s = \frac{\hat{x}_s - z_s}{1 - t_s}, \qquad v_s = \frac{x_s - z_s}{1 - t_s}.
\label{eq:velocity}
\end{equation}
For the instance stream, the random start of \cref{eq:interp} is unnecessary: $C$ has no fixed target to reach, and a deterministic map needs no sampling.
We fix its start at $z_\text{i} = 0$ and its timestep at $t_\text{i} = 0$, the inference timestep of all streams.
The stream regresses its prediction $\hat{x}_\text{i}$ of $C$ from image features in one step.

\paragraph{Stream Mask.}
Each stream embeds its input map into $L$ patch tokens, concatenates them with the frozen DINOv3 features~\cite{dinov3} of the same patches and projects them to tokens $X_s$.
The three streams form one sequence $X = [X_\text{p}; X_\text{n}; X_\text{i}]$ of $3L$ tokens, which passes through DiT blocks~\cite{dit} with adaLN-Zero conditioned on the timestep of each stream.
Each stream decodes its output tokens with its own linear layer.
Let $s_j$ denote the stream of token $j$, and let $q_j$ and $k_l$ be the query of token $j$ and the key of token $l$.
Every attention layer adds a stream mask $M_{jl}$ to the logits as
\begin{equation}
a_{jl} = \operatorname*{softmax}_l \big( \ell_{jl} \big), \qquad \ell_{jl} = \frac{q_j^\top k_l}{\sqrt{d}} + M_{jl},
\label{eq:attn}
\end{equation}
where $d$ is the head dimension and
\begin{equation}
M_{jl} =
\begin{cases}
0, & s_j = s_l \ \text{or} \ s_j, s_l \in \{\text{p}, \text{n}\}, \\
-\infty, & \text{otherwise}.
\end{cases}
\label{eq:mask}
\end{equation}

The point map gives the position of every surface, and the normals give its orientation.
The two geometry streams thus attend to each other.
The mask blocks attention between the geometry and instance streams in both directions.
The instance map thus does not depend on the noise of the geometry streams during training and stays a deterministic function of the image.

\begin{algorithm}[t]
\caption{One training step of PINE}
\label{alg:pine}
\begin{algorithmic}[1]
\REQUIRE image $I$; intrinsics $K$ or $\varnothing$; targets $x_\text{p}$, $x_\text{n}$ and instance ids; reference intrinsics $K_\text{ref}$; weights $\theta$
\ENSURE updated weights $\theta$
\STATE $F \leftarrow \text{DINOv3}(I)$ \hfill $\triangleright$ frozen
\IF{$K = \varnothing$}
\STATE $K \leftarrow$ default camera, $e \leftarrow$ learned embedding
\ELSE
\STATE $e \leftarrow 0$
\ENDIF
\STATE $\rho_j \leftarrow$ ray position of every patch $j$ \hfill $\triangleright$ \cref{eq:position}
\FOR{$s \in \{\text{p}, \text{n}\}$}
\STATE sample $t_s \in [0, 1]$ and $\epsilon_s \sim \mathcal{N}(0, \mathbf{I})$
\STATE $z_s \leftarrow t_s\, x_s + (1 - t_s)\, \sigma \epsilon_s$ \hfill $\triangleright$ \cref{eq:interp}
\ENDFOR
\STATE $z_\text{i} \leftarrow 0$, $t_\text{i} \leftarrow 0$
\STATE $X_s \leftarrow \text{Embed}_s(z_s, F, \rho)$ for $s \in \{\text{p}, \text{n}, \text{i}\}$
\STATE $X \leftarrow [X_\text{p}; X_\text{n}; X_\text{i}]$
\STATE $g_j \leftarrow \text{Linear}(\text{LayerNorm}(F_j))$ for every patch $j$
\FOR{every DiT block and head}
\STATE rotate $q_j$ and $k_l$ by $\rho_j$ and $\rho_l$
\STATE $\ell_{jl} \leftarrow q_j^\top k_l / \sqrt{d} + M_{jl}$ \hfill $\triangleright$ \cref{eq:attn,eq:mask}
\IF{$s_j = s_l \in \{\text{n}, \text{i}\}$}
\STATE $\ell_{jl} \leftarrow \ell_{jl} + (U g_j)^\top (V g_l)$ \hfill $\triangleright$ \cref{eq:logit}
\ENDIF
\STATE $X \leftarrow \text{Block}\big(X, \operatorname{softmax}_l(\ell_{jl}), t_s, e\big)$
\ENDFOR
\STATE $\hat{x}_s \leftarrow \text{Linear}_s(X_s)$ for $s \in \{\text{p}, \text{n}, \text{i}\}$
\STATE update $\theta$ with $\nabla_\theta \mathcal{L}$ \hfill $\triangleright$ \cref{eq:loss}
\end{algorithmic}
\textbf{Inference:} set $t_s = 0$ and $z_s = \sigma \epsilon_s$ in lines 9 and 10, skip line 25 and return $(P, N, C) = (\hat{x}_\text{p}, \hat{x}_\text{n}, \hat{x}_\text{i})$.
If $K = \varnothing$, recover the focal length by \cref{eq:selfcalib}.
\end{algorithm}

\subsection{Camera-Aware Ray Encoding}
\label{sec:ray_encoding}

\begin{figure}[t]
\centering
\includegraphics[width=\linewidth]{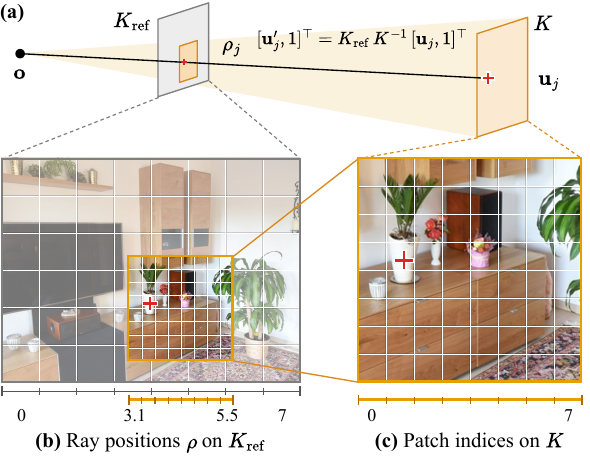}
\caption{\textbf{Camera-aware ray encoding.}
(a) The position $\rho_j$ of token $j$ is the point where its ray meets the image plane of a reference camera.
(b) With ray positions, a zoomed crop lands inside the full view, and the point at the red cross keeps one position.
(c) With patch indices, the crop has its own grid and gives the same point a second position.}
\label{fig:ray_encoding}
\end{figure}

A point map back-projects every pixel $\mathbf{u}$ along its camera ray, $P(\mathbf{u}) = Z(\mathbf{u})\, K^{-1} [\mathbf{u}, 1]^\top$ with depth $Z$.
The intrinsics thus fix the direction along which each depth is placed.
PINE therefore positions every token by the ray of its patch instead of its pixel.

\paragraph{Ray Positions.}
The ray through the center $\mathbf{u}_j$ of patch $j$ meets the image plane of a reference camera with intrinsics $K_\text{ref}$ at the pixel $\mathbf{u}'_j$, given by
\begin{equation}
[\mathbf{u}'_j, 1]^\top = K_\text{ref} K^{-1} [\mathbf{u}_j, 1]^\top.
\label{eq:position}
\end{equation}
The position of token $j$ is this pixel in patch units, $\rho_j = \mathbf{u}'_j / p - \tfrac{1}{2}$, where $p$ is the patch size.
When $K = K_\text{ref}$, $\rho_j$ is the integer patch index of token $j$.
Patch indices are thus the special case in which every image comes from one fixed camera.
The positions replace the patch indices in the sine-cosine encodings that form $X_s$ and in the 2D rotary encoding of the queries and keys.
Since the rotary encoding makes the positional part of $q_j^\top k_l$ a function of $\rho_l - \rho_j$, attention sees the offset between two rays rather than their distance in pixels.
\Cref{fig:ray_encoding}a shows this construction.
With ray positions in \cref{fig:ray_encoding}b, a zoomed crop lands inside the full view, and the point at the red cross keeps one position.
With patch indices in \cref{fig:ray_encoding}c, the crop has its own grid and gives the same point a second position.

\paragraph{Unknown Intrinsics.}
Without intrinsics, PINE uses a default camera with the field of view of the reference camera and its principal point at the image center.
This camera places the tokens on a regular grid like patch indices.
Given that these positions carry no information about the true intrinsics, a learned embedding added to the timestep condition of every stream marks the missing intrinsics.
PINE trains on both cases, and one model serves images whether or not intrinsics are given.
After prediction, PINE recovers the focal length from its own point map, as MoGe does~\cite{moge}.
With square pixels and the principal point at the image center, the shifted point map lies on the camera rays,
\begin{equation}
K_f \big( P(\mathbf{u}) + \Delta \big) \propto [\mathbf{u}, 1]^\top \quad \text{for every pixel } \mathbf{u},
\label{eq:selfcalib}
\end{equation}
where $K_f$ is the intrinsic matrix with focal length $f$ and $\Delta \in \mathbb{R}^3$ is the unknown shift of the affine-invariant point map.
For a fixed $\Delta_z$, \cref{eq:selfcalib} is linear in $1/f$, $\Delta_x$ and $\Delta_y$, which we solve by least squares.
A one-dimensional search over $\Delta_z$ completes the solution.

\subsection{Grounding Attention}
\label{sec:grounding}

\begin{figure}[t]
\centering
\includegraphics[width=0.95\linewidth]{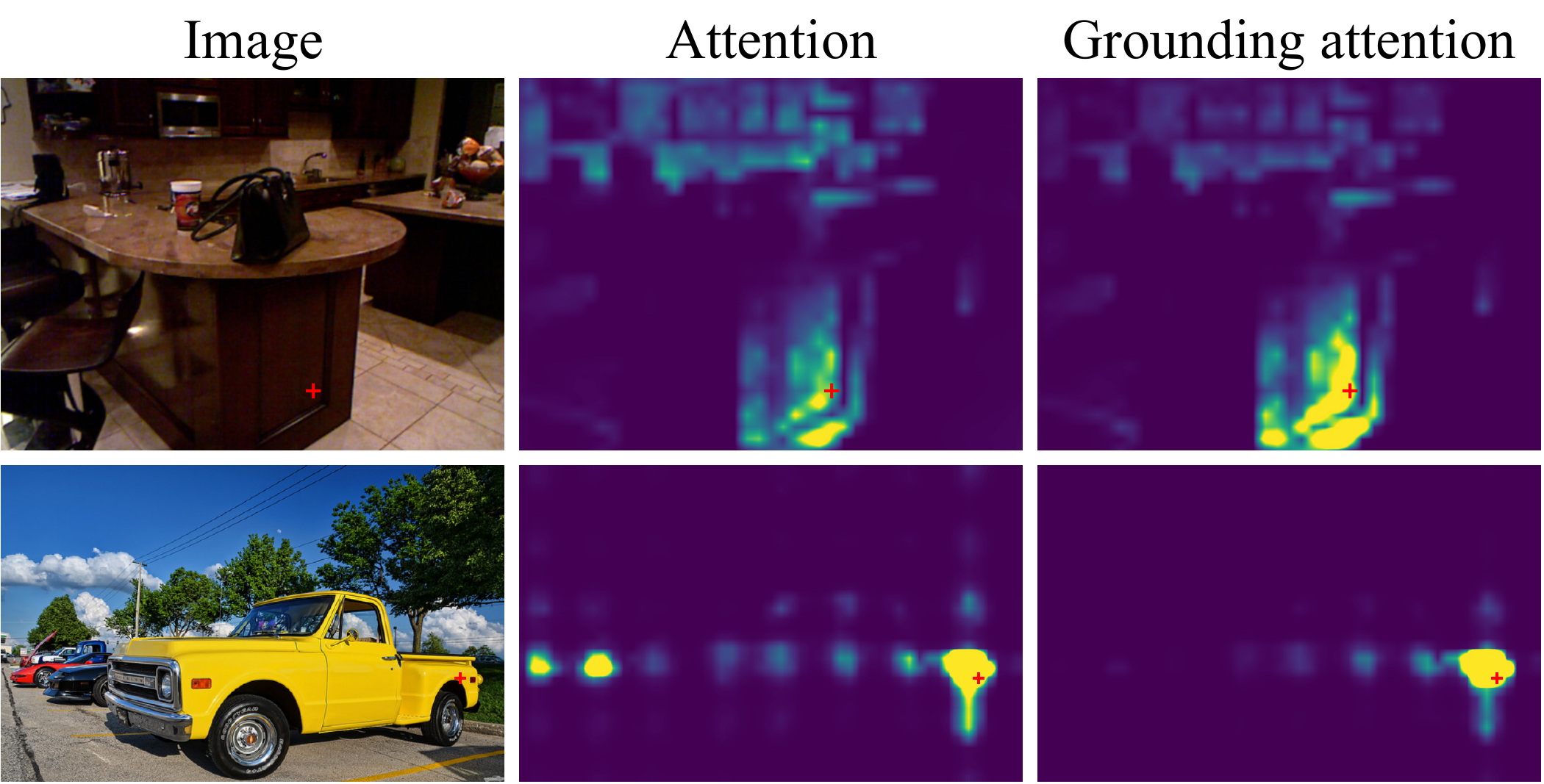}
\caption{\textbf{Grounding attention.}
Each row shows an image and the attention of the query patch at the red cross, without and with grounding attention.
Grounding attention keeps the attention on the region of the query patch.}
\label{fig:grounding}
\end{figure}

A normal is constant over a planar surface, and an instance label is constant over an object.
Both maps thus depend on how pixels group into surfaces and objects.
Self-supervised ViT features already encode this grouping in their affinities~\cite{tokencut,DeepSpectralMethods}.
Yet in \cref{eq:attn}, the DINOv3 features enter only the tokens, and attention infers from them which tokens should exchange information.
PINE therefore writes the affinity into the attention logits as a bias.
A layer normalization and a linear projection map the DINOv3 feature of patch $j$ to a grounding code $g_j$.
For normal-normal and instance-instance pairs, every layer adds a bias to $\ell_{jl}$ as
\begin{equation}
b_{jl} = (U g_j)^\top (V g_l),
\label{eq:logit}
\end{equation}
where each head of each block has its own learned low-rank projections $U$ and $V$.
The bias raises the attention between patches that look alike regardless of their positions and of the current geometry estimate.
\Cref{fig:grounding} shows that grounding attention moves the attention of a patch onto its own object.
The point stream does not use grounding attention, because depth also depends on the global layout: two walls that look alike can lie at different depths.

\subsection{Training}
\label{sec:objective}

\paragraph{Training Losses.}
We train PINE end-to-end with
\begin{equation}
\mathcal{L} = \mathcal{L}_\text{vel} + \lambda_\text{rel} \mathcal{L}_\text{rel} + \lambda_\text{grad} \mathcal{L}_\text{grad} + \mathcal{L}_\text{sink} + \lambda_\text{inst} \mathcal{L}_\text{inst},
\label{eq:loss}
\end{equation}
where $\lambda_\text{rel}$, $\lambda_\text{grad}$ and $\lambda_\text{inst}$ weight the terms.
Let $\Omega$ denote the valid non-sky pixels of a target and $\langle \cdot \rangle_\Omega$ the mean over them.
The velocity loss supervises both streams via \cref{eq:velocity},
\begin{equation}
\mathcal{L}_\text{vel} = \sum_{s \in \{\text{p}, \text{n}\}} \big\langle \|\hat{v}_s - v_s\|^2 \big\rangle_\Omega .
\label{eq:loss_v}
\end{equation}
The other losses act on the clean predictions.
The relative point loss $\mathcal{L}_\text{rel} = \big\langle \|\hat{x}_\text{p} - x_\text{p}\|_1 / \|x_\text{p}\| \big\rangle_\Omega$ divides the error of every point by its distance to the center.
The gradient loss $\mathcal{L}_\text{grad}$ follows MiDaS~\cite{midas} and sharpens depth edges.
The normal loss $\mathcal{L}_\text{sink}$ follows Marigold V2~\cite{marigoldv2}, and the instance loss $\mathcal{L}_\text{inst}$ follows gen2seg~\cite{gen2seg}, as detailed in \cref{sec:supp_impl}.

\section{Experiments}
\label{sec:exp}

\subsection{Experimental Setups}
\label{sec:exp_setup}

\paragraph{Implementation Details.}
PINE starts from the pretrained weights of PointDiT~\cite{pointdit} and trains on the 139K synthetic and 24K real images of \cref{tab:supp_data}.
Its B, L and H models use a frozen DINOv3 ViT-B/16, ViT-L/16 and ViT-H+/16~\cite{dinov3}.
The reference camera has the focal length $f_\text{ref} = 420$ pixels and its principal point at the image center.
We clamp the denominator $1 - t_s$ of \cref{eq:velocity} at 0.05.
We crop each training image at a random aspect ratio in $[0.5, 3.5]$, flip it and jitter its color, and \cref{sec:supp_impl} lists the full augmentation.
The loss weights are $\lambda_\text{rel} = 0.1$, $\lambda_\text{grad} = 0.5$ and $\lambda_\text{inst} = 0.02$, and the sky pixels add the error of \cref{eq:loss_v} with a weight of 0.01.
We train for 20 epochs with AdamW, a constant learning rate of $10^{-5}$ and a batch size of 64 in bfloat16.
Training PINE-L takes about 20 hours on 8 A100 GPUs.

\paragraph{Evaluation.}
We evaluate point maps and depth on 3{,}444 images from DIODE~\cite{diode}, KITTI~\cite{kitti}, NYUv2~\cite{nyuv2}, ETH3D~\cite{eth3d}, HAMMER~\cite{hammer}, iBims-1~\cite{ibims} and Booster~\cite{booster}, and we report the relative error Rel and the accuracy $\delta_1$.
We evaluate normals on NYUv2~\cite{nyuv2}, ScanNet~\cite{scannet}, iBims-1~\cite{ibims} and Sintel~\cite{sintel} with the protocol of DSINE~\cite{dsine}, and we report the mean angular error and the accuracy within $11.25^\circ$.
We evaluate instances on \cocoexc~\cite{coco}, DRAM~\cite{dram}, EgoHOS~\cite{egohos}, iShape~\cite{ishape} and PIDRay~\cite{pidray}, where \cocoexc holds the 46 COCO categories that gen2seg~\cite{gen2seg} excludes from training.
Each method takes one point prompt per instance, and we report the mIoU of the resulting masks.
\Cref{sec:supp_eval} gives the alignment, the metric definitions and the prompt protocol.

\begin{table}[t]
\centering
\caption{\textbf{Average results on point maps and depth} over the 3{,}444 images of the seven datasets.
Params and Latency are from \cref{tab:cost_single}, and the latency of PINE includes its normals and instances.
Best results are in bold, separately for the settings with and without intrinsics.}
\footnotesize
\setlength{\tabcolsep}{3pt}
\renewcommand{\arraystretch}{0.92}
\begin{tabular}{@{}lcccccc@{}}
    \toprule
    \multirow{2}{*}{Method} & \multicolumn{2}{c}{Point map} & \multicolumn{2}{c}{Depth} & \multirow{2}{*}{\begin{tabular}[c]{@{}c@{}}Params\\(B)\end{tabular}} & \multirow{2}{*}{\begin{tabular}[c]{@{}c@{}}Latency\\(ms)\end{tabular}} \\
    \cmidrule(lr){2-3} \cmidrule(lr){4-5}
    & Rel$^\text{p}$ $\downarrow$ & $\delta_1^\text{p}$ $\uparrow$ & Rel$^\text{d}$ $\downarrow$ & $\delta_1^\text{d}$ $\uparrow$ & & \\
    \midrule
    \rowcolor{gray!15}\multicolumn{7}{l}{\textbf{\textit{Without intrinsics}}} \\
    GeometryCrafter~\cite{geometrycrafter} & 5.45 & 96.75 & 3.52 & 97.84 & 2.67 & 964.7 \\
    PPD~\cite{ppd} & 5.54 & 96.59 & 3.88 & 97.78 & 0.80 & 691.9 \\
    Depth Pro~\cite{depthpro} & 5.71 & 96.71 & 3.72 & 97.74 & 0.95 & 909.3 \\
    UniDepthV2~\cite{unidepthv2} & 4.45 & 97.35 & 2.86 & 98.52 & 0.35 & 23.3 \\
    DA3~\cite{da3} & 4.77 & 96.63 & 3.22 & 97.81 & 1.36 & 97.3 \\
    MoGe~\cite{moge} & 4.21 & 97.45 & 3.10 & 98.01 & 0.31 & 86.0 \\
    MoGe-2~\cite{moge2} & 4.53 & 97.46 & 2.87 & 98.50 & 0.33 & 67.9 \\
    FoundationGeo~\cite{foundationgeo} & 4.57 & 97.71 & 2.95 & 98.52 & 0.31 & 90.1 \\
    MoGe-3~\cite{moge3} & 4.79 & 97.36 & 2.90 & 98.43 & 0.37 & 360.5 \\
    Marigold V2~\cite{marigoldv2} & -- & -- & 3.55 & \textbf{98.59} & 21.4 & 491.6 \\
    \midrule
    PointDiT-B~\cite{pointdit} & 5.84 & 96.69 & 3.70 & 97.84 & 0.22 & 23.3 \\
    PointDiT-L~\cite{pointdit} & 4.90 & 97.42 & 3.15 & 98.22 & 0.77 & 45.8 \\
    PointDiT-H~\cite{pointdit} & 4.45 & 97.93 & 2.81 & 98.51 & 1.81 & 62.5 \\
    \midrule
    PINE-B & 5.62 & 96.75 & 3.70 & 97.75 & 0.24 & 43.8 \\
    PINE-L & 4.70 & 97.78 & 3.04 & 98.29 & 0.80 & 82.6 \\
    PINE-H & \textbf{4.16} & \textbf{98.13} & \textbf{2.79} & 98.51 & 1.85 & 136.6 \\
    \midrule
    \rowcolor{gray!15}\multicolumn{7}{l}{\textbf{\textit{With intrinsics}}} \\
    Depth Pro~\cite{depthpro} & 4.53 & 97.35 & 3.72 & 97.74 & 0.95 & 909.3 \\
    UniDepthV2~\cite{unidepthv2} & 3.70 & 97.95 & 2.87 & 98.52 & 0.35 & 23.3 \\
    DA3~\cite{da3} & 3.96 & 97.10 & 3.17 & 97.89 & 1.36 & 97.3 \\
    MoGe-2~\cite{moge2} & 4.38 & 97.58 & 2.87 & 98.50 & 0.33 & 67.9 \\
    MoGe-3~\cite{moge3} & 4.70 & 97.35 & 2.90 & 98.43 & 0.37 & 360.5 \\
    FoundationGeo~\cite{foundationgeo} & 4.40 & 97.85 & 2.95 & 98.52 & 0.31 & 90.1 \\
    \midrule
    PINE-B & 4.69 & 97.06 & 3.60 & 97.75 & 0.24 & 43.8 \\
    PINE-L & 3.76 & 97.90 & 2.92 & 98.32 & 0.80 & 82.6 \\
    PINE-H & \textbf{3.33} & \textbf{98.21} & \textbf{2.65} & \textbf{98.54} & 1.85 & 136.6 \\
    \bottomrule
\end{tabular}
\label{tab:comparison}
\end{table}

\subsection{Main Results}
\label{sec:exp_results}

\begin{table}[t]
\centering
\caption{\textbf{Quantitative comparison on zero-shot surface normal estimation.}
Mean angular error in degrees and accuracy within $11.25^\circ$ under the full-frame protocol of DSINE.
Best results are in bold, separately for the settings with and without intrinsics.}
\setlength{\tabcolsep}{2pt}
\resizebox{\columnwidth}{!}{
\begin{tabular}{lccccccccc}
    \toprule
    \multirow{2}{*}{Method} & \multicolumn{2}{c}{NYUv2} & \multicolumn{2}{c}{ScanNet} & \multicolumn{2}{c}{iBims-1} & \multicolumn{2}{c}{Sintel} & \multirow{2}{*}{\begin{tabular}[c]{@{}c@{}}Avg.\\Rank $\downarrow$\end{tabular}} \\
    \cmidrule(lr){2-3} \cmidrule(lr){4-5} \cmidrule(lr){6-7} \cmidrule(lr){8-9}
    & Mean$\downarrow$ & $11.25^\circ\uparrow$ & Mean$\downarrow$ & $11.25^\circ\uparrow$ & Mean$\downarrow$ & $11.25^\circ\uparrow$ & Mean$\downarrow$ & $11.25^\circ\uparrow$ & \\
    \midrule
    \rowcolor{gray!15}\multicolumn{10}{l}{\textbf{\textit{Without intrinsics}}} \\
    GeoWizard~\cite{geowizard} & 18.9 & 50.7 & 17.4 & 53.8 & 19.3 & 63.0 & 40.3 & 12.3 & 12.0 \\
    StableNormal~\cite{stablenormal} & 18.6 & 53.5 & 17.1 & 57.4 & 18.2 & 65.0 & 36.7 & 14.1 & 11.0 \\
    DSINE~\cite{dsine} & 16.4 & 59.6 & 16.2 & 61.0 & 17.1 & 67.4 & 34.9 & 21.5 & 9.3 \\
    Lotus-D~\cite{lotus} & 16.2 & 59.8 & 14.7 & 64.0 & 17.1 & 66.4 & 32.3 & 22.4 & 7.8 \\
    E2E-FT~\cite{e2eft} & 16.5 & 60.4 & 14.7 & 66.1 & 16.1 & 69.7 & 33.5 & 22.3 & 7.6 \\
    FE2E~\cite{fe2e} & 16.2 & 59.6 & 13.8 & 67.2 & 15.1 & 70.6 & 31.2 & 22.3 & 5.5 \\
    Marigold V2~\cite{marigoldv2} & 16.0 & 61.6 & 13.8 & 67.9 & 15.2 & \textbf{71.9} & 29.7 & 27.3 & 4.3 \\
    TransNormal-2~\cite{transnormal2} & 14.9 & 62.4 & 12.9 & 69.1 & 15.1 & 70.1 & 28.2 & 28.1 & 2.9 \\
    MoGe-3~\cite{moge3} & \textbf{14.7} & 62.3 & \textbf{12.8} & 68.9 & \textbf{14.8} & 70.1 & 26.4 & 29.1 & 2.4 \\
    \midrule
    PINE-B & 16.4 & 59.2 & 15.3 & 63.0 & 16.5 & 68.5 & 30.4 & 24.1 & 8.0 \\
    PINE-L & 15.7 & 61.9 & 14.1 & 66.4 & 15.5 & 70.9 & 27.5 & 29.6 & 4.3 \\
    PINE-H & 15.3 & \textbf{63.1} & 13.3 & \textbf{69.2} & 15.1 & 71.8 & \textbf{25.6} & \textbf{32.5} & \textbf{1.8} \\
    \midrule
    \rowcolor{gray!15}\multicolumn{10}{l}{\textbf{\textit{With intrinsics}}} \\
    PINE-B & 16.3 & 59.7 & 14.9 & 64.7 & 16.2 & 69.0 & 30.0 & 24.6 & -- \\
    PINE-L & 15.6 & 62.5 & 13.6 & 68.5 & 15.1 & 71.7 & 26.5 & 30.9 & -- \\
    PINE-H & \textbf{15.1} & \textbf{63.5} & \textbf{13.0} & \textbf{70.0} & \textbf{14.8} & \textbf{72.2} & \textbf{25.0} & \textbf{34.0} & -- \\
    \bottomrule
\end{tabular}
}
\label{tab:normal_comparison}
\end{table}

\begin{table}[t]
\centering
\caption{\textbf{Point-prompted instance segmentation.}
mIoU with one point prompt per instance, where L, M and S split \cocoexc into large, medium and small instances.
$\ddagger$: re-evaluated under our protocol; SimpleClick is taken from gen2seg, and $\dagger$ merges the instances of one class.}
\setlength{\tabcolsep}{3pt}
\resizebox{\columnwidth}{!}{
\begin{tabular}{lccccccc}
\toprule
Method & COCO$^\text{L}_\text{exc}$ & COCO$^\text{M}_\text{exc}$ & COCO$^\text{S}_\text{exc}$ & DRAM & EgoHOS & iShape & PIDRay \\
\midrule
SAM (ViT-H)~\cite{sam}$^\ddagger$ & 58.7 & \textbf{60.5} & \textbf{57.2} & 49.8 & \textbf{58.2} & \textbf{40.9} & 44.2 \\
SimpleClick~\cite{simpleclick} & 1.4 & 0.6 & 0.2 & 2.4 & 1.6 & 1.6$^\dagger$ & 1.5 \\
gen2seg~\cite{gen2seg}$^\ddagger$ & 59.6 & 38.7 & 8.3 & 46.6 & 42.9 & 9.9 & 35.6 \\
\midrule
PINE-B & 68.6 & 48.5 & 9.2 & 54.2 & 54.1 & 15.1 & 41.8 \\
PINE-L & 75.2 & 52.1 & 9.3 & 58.4 & 52.3 & 18.4 & \textbf{45.6} \\
PINE-H & \textbf{76.3} & 51.8 & 9.2 & \textbf{60.5} & 53.7 & 15.4 & 42.5 \\
\bottomrule
\end{tabular}
}
\label{tab:instance_comparison}
\end{table}

\begin{figure*}[t]
\centering
\includegraphics[width=\linewidth]{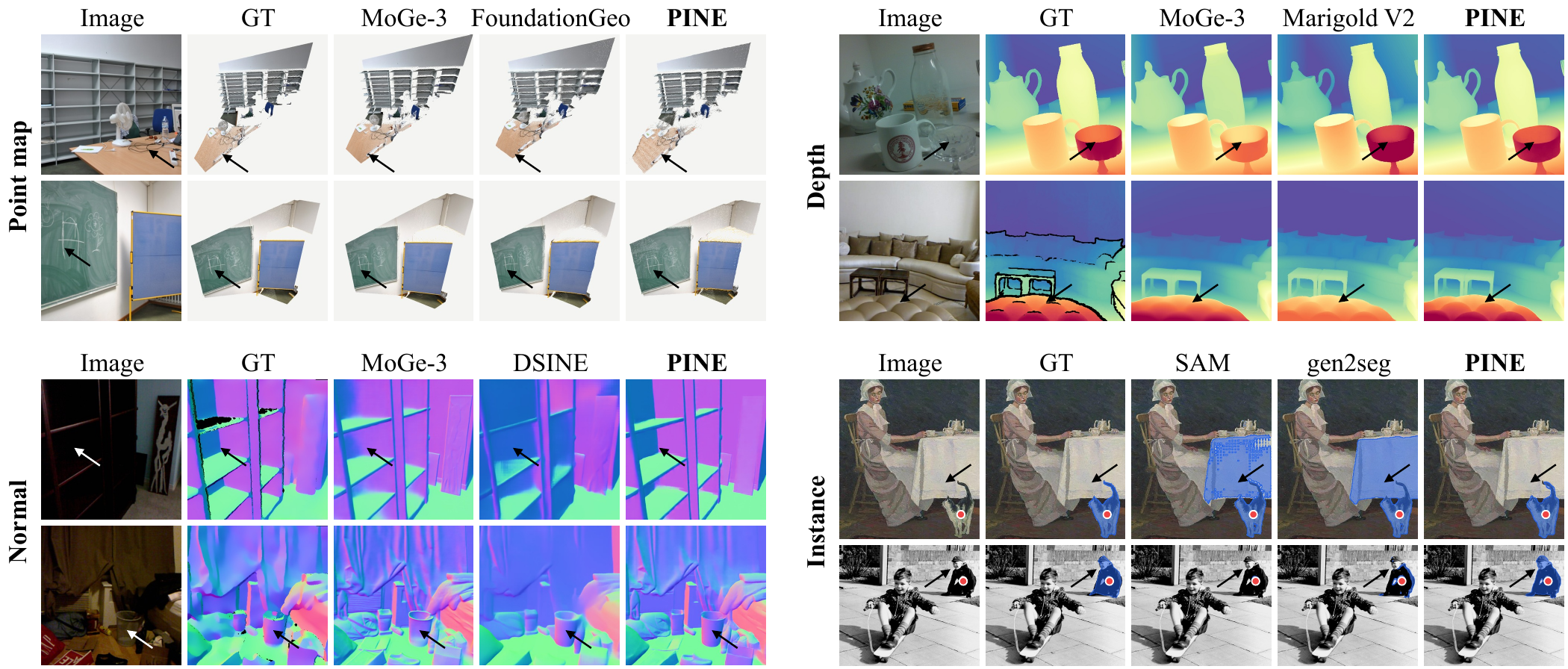}
\caption{\textbf{Qualitative comparison on point maps, depth, normals and instances.}
The red dot marks the point prompt of each instance.
Arrows highlight the regions we significantly improved.}
\label{fig:qualitative}
\end{figure*}

\begin{table}[t]
\centering
\caption{\textbf{Field-of-view estimation.}
Mean absolute error in degrees of the horizontal field of view recovered from the predicted point map without intrinsics.
The Avg column is the sample-weighted mean over all $3{,}444$ images.}
\setlength{\tabcolsep}{3pt}
\resizebox{\columnwidth}{!}{
\begin{tabular}{lcccccccc}
\toprule
Method & DIODE & KITTI & NYUv2 & ETH3D & HAMMER & iBims-1 & Booster & Avg \\
\midrule
MoGe-2~\cite{moge2} & \textbf{6.61} & 10.64 & \textbf{1.60} & \textbf{3.73} & 12.18 & 4.77 & 9.88 & 7.28 \\
MoGe-3~\cite{moge3} & 7.70 & 12.02 & 1.61 & 4.00 & 13.75 & 5.54 & 11.78 & 8.22 \\
FoundationGeo~\cite{foundationgeo} & 8.01 & \textbf{4.56} & 2.57 & 4.43 & 12.18 & 4.49 & 8.68 & 6.70 \\
\midrule
PointDiT-B~\cite{pointdit} & 8.11 & 17.65 & 5.54 & 20.11 & 17.48 & 6.70 & 8.12 & 13.08 \\
PointDiT-L~\cite{pointdit} & 8.34 & 14.02 & 4.48 & 11.00 & 13.16 & 5.35 & 8.48 & 10.03 \\
PointDiT-H~\cite{pointdit} & 9.38 & 19.70 & 6.88 & 12.92 & 19.30 & 6.81 & 7.26 & 13.46 \\
\midrule
PINE-B & 7.97 & 5.76 & 7.30 & 9.90 & 6.66 & 6.26 & 7.59 & 7.33 \\
PINE-L & 7.34 & 5.45 & 7.69 & 7.80 & 4.97 & 4.93 & \textbf{5.16} & 6.48 \\
PINE-H & 7.76 & 4.91 & 4.01 & 6.97 & \textbf{3.70} & \textbf{3.86} & 5.63 & \textbf{5.35} \\
\bottomrule
\end{tabular}
}
\label{tab:fov_main}
\end{table}
\paragraph{Results on Point Maps and Depth.}
PINE-H has the lowest average $\text{Rel}^\text{p}$ and the highest average $\delta_1^\text{p}$ of all methods in \cref{tab:comparison} in both settings, although it also predicts normals and instances in the same forward pass.
Without intrinsics, its largest gain is on HAMMER, where it lowers $\text{Rel}^\text{p}$ from 3.51 for PointDiT-H, the second-best method, to 2.96, by 16\%.
With intrinsics, it lowers the average $\text{Rel}^\text{p}$ to 3.33, against 3.70 for UniDepthV2.
At every model size, PINE has a lower $\text{Rel}^\text{p}$ than PointDiT.
This gain does not come from the additional training: PointDiT with our recipe reaches only 5.01 against 4.82 for PINE in \cref{tab:ablation_components}.
This shows the benefit of positioning tokens by camera rays rather than by pixels.
The gain carries over to depth, where PINE-H has the lowest average $\text{Rel}^\text{d}$ in both settings, 2.79 without and 2.65 with intrinsics.
\Cref{fig:qualitative} compares the 3D shapes of PINE-H and the baselines.

\paragraph{Results on Surface Normals.}
PINE-H has the best average rank of all methods in \cref{tab:normal_comparison}.
Its largest gain is on Sintel, where it reaches a mean angular error of $25.6^\circ$ against $26.4^\circ$ for MoGe-3, the second-best method.
It also raises the Sintel accuracy within $11.25^\circ$ from 29.1\% to 32.5\%.
PINE learns normals from only 139K synthetic images, while MoGe-3 trains on more than 8.9M frames.
With intrinsics, PINE-H lowers its error further on all four datasets.
\Cref{fig:qualitative} shows the normals of PINE-H and baselines.

\paragraph{Results on Instances.}
PINE-H exceeds SAM on large COCO objects and DRAM by 17.6 and 10.7 mIoU in \cref{tab:instance_comparison}.
It also outperforms gen2seg, a diffusion-based instance model, on every benchmark.
PINE learns instances from only 87K images, while SAM trains on 11M images with 1.1B masks.
This shows that instances can be learned as an additional stream of a geometry model, without a large mask corpus.
On small COCO objects, PINE-H trails SAM with 9.2 against 57.2 mIoU, since a small object covers only a few patches.
\Cref{fig:qualitative} shows that PINE-H gives more complete masks of large objects than gen2seg.

\paragraph{Results on Camera and Planes.}
PINE-H recovers the field of view with the lowest average error in \cref{tab:fov_main}, $5.35^\circ$ against $6.70^\circ$ for FoundationGeo and $13.46^\circ$ for PointDiT-H.
Its largest gain is on HAMMER, where PINE-H reaches $3.70^\circ$ against $19.30^\circ$ for PointDiT-H and $12.18^\circ$ for MoGe-2.
At every model size, PINE lowers the field-of-view error of PointDiT, by up to 60\% for PINE-H.
The camera-aware ray encoding thus lets PINE recover the intrinsics.
On iBims-1, PINE-H lowers the planarity error of PointDiT-H from 0.62 to 0.54\,cm in \cref{tab:ibims_planarity}.

\begin{figure}[t]
\centering
\includegraphics[width=\linewidth]{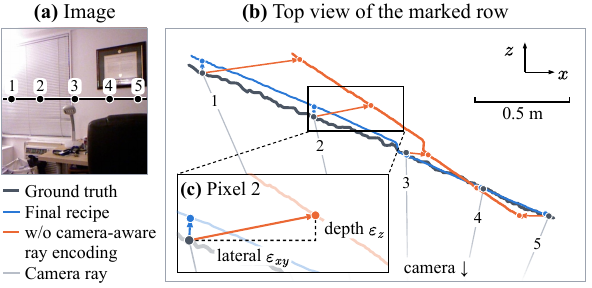}
\caption{\textbf{Lateral and depth error with and without the camera-aware ray encoding.} (a) Input image with five marked pixels on one row. (b) Top view of the points of this row. Each arrow points from the ground-truth point of a marked pixel to its predicted point. (c) The error of pixel 2, split into a lateral part $\varepsilon_{xy}$ and a depth part $\varepsilon_z$.}
\label{fig:ablation_camera_mechanism}
\end{figure}

\subsection{Ablation Study}
\label{sec:ablation}
\Cref{tab:ablation_components} ablates the three components of PINE: the camera-aware ray encoding, the stream mask and grounding attention.
To limit the cost, we train each variant as a PINE-L model on Hypersim and Virtual KITTI~2 only, and the other settings follow \cref{sec:exp_setup}.
To separate the gain of PINE from the additional training, we also train the point-only PointDiT architecture with the same recipe.
The point maps, the depth and the normals use the benchmarks of \cref{sec:exp_setup}, and the instances use \cocoexc and DRAM.

\begin{table*}[t]
\centering
\caption{\textbf{Ablation of the camera-aware ray encoding, the stream mask and grounding attention.}
The first row is the final recipe with the mean and standard deviation over three seeds.
PE$_\text{img}$ and PE$_\text{geo}$ are sine-cosine positions on the image features and the geometry tokens, and X$\leftarrow$Y means that the tokens of X attend to the tokens of Y.}
\setlength{\tabcolsep}{4pt}
\resizebox{\textwidth}{!}{
\begin{tabular}{ccccccccccccccc}
\toprule
\multicolumn{3}{c}{Camera-aware ray encoding} & \multicolumn{3}{c}{Stream mask} & \multirow{2}{*}{Grounding} & \multicolumn{2}{c}{Rel$^\text{p}$ $\downarrow$} & \multicolumn{2}{c}{Rel$^\text{d}$ $\downarrow$} & \multicolumn{2}{c}{Normal $\downarrow$} & \multicolumn{2}{c}{Instance $\uparrow$} \\
\cmidrule(lr){1-3} \cmidrule(lr){4-6} \cmidrule(lr){8-9} \cmidrule(lr){10-11} \cmidrule(lr){12-13} \cmidrule(lr){14-15}
RoPE & PE$_\text{img}$ & PE$_\text{geo}$ & N$\leftarrow$P & I$\leftarrow$G & G$\leftarrow$I & & w/o $K$ & with $K$ & w/o $K$ & with $K$ & w/o $K$ & with $K$ & \cocoexc & DRAM \\
\midrule
\cmark & \cmark & \cmark & \cmark & \xmark & \xmark & Attn. & \textbf{4.82{\scriptsize$\pm$.03}} & \textbf{3.94{\scriptsize$\pm$.03}} & \textbf{3.15{\scriptsize$\pm$.01}} & \textbf{3.05{\scriptsize$\pm$.02}} & \textbf{18.23{\scriptsize$\pm$.03}} & \textbf{17.81{\scriptsize$\pm$.02}} & \textbf{33.0{\scriptsize$\pm$.05}} & \textbf{54.8{\scriptsize$\pm$.28}} \\
\midrule
\xmark & \xmark & \xmark & \cmark & \xmark & \xmark & Attn. & 5.05 & 5.05 & 3.14 & 3.14 & 18.23 & 18.23 & 33.1 & 52.3 \\
\cmark & \xmark & \xmark & \cmark & \xmark & \xmark & Attn. & 5.05 & 4.69 & 3.21 & 3.10 & 18.33 & 18.18 & 33.1 & 55.0 \\
\cmark & \cmark & \xmark & \cmark & \xmark & \xmark & Attn. & 5.04 & 4.77 & 3.20 & 3.02 & 18.22 & 17.84 & 32.6 & 53.3 \\
\midrule
\cmark & \cmark & \cmark & \xmark & \xmark & \xmark & Attn. & 4.81 & 3.91 & 3.16 & 3.03 & 18.23 & 17.94 & 33.0 & 53.9 \\
\cmark & \cmark & \cmark & \cmark & \cmark & \xmark & Attn. & 4.93 & 3.97 & 3.18 & 3.06 & 18.25 & 17.80 & 32.4 & 55.3 \\
\cmark & \cmark & \cmark & \cmark & \cmark & \cmark & Attn. & 4.96 & 3.93 & 3.15 & 3.04 & 18.29 & 17.83 & 33.0 & 53.4 \\
\midrule
\cmark & \cmark & \cmark & \cmark & \xmark & \xmark & \xmark & 4.79 & 3.96 & 3.14 & 3.05 & 18.47 & 17.92 & 32.1 & 52.5 \\
\cmark & \cmark & \cmark & \cmark & \xmark & \xmark & adaLN & 4.85 & 3.90 & 3.13 & 3.01 & 18.32 & 17.90 & 33.1 & 52.3 \\
\midrule

\multicolumn{7}{l}{PointDiT~\cite{pointdit} (our recipe)} & 5.01 & -- & 3.20 & -- & -- & -- & -- & -- \\
\bottomrule
\end{tabular}
}
\label{tab:ablation_components}
\end{table*}

\paragraph{Camera-Aware Ray Encoding.}
In \cref{tab:ablation_components}, without the encoding, intrinsics have no effect, and $\text{Rel}^\text{p}$ stays at 5.05 in both settings.
With the encoding, intrinsics reduce $\text{Rel}^\text{p}$ from 4.82 to 3.94 and the mean angular error from 18.23 to 17.81.
Without intrinsics, the encoding still reduces $\text{Rel}^\text{p}$ from 5.05 to 4.82.
We split the error of each point into a lateral part $\varepsilon_{xy}$ and a depth part $\varepsilon_z$.
Averaged over the valid pixels with intrinsics, the encoding reduces $\varepsilon_{xy}$ from 2.28 to 1.27, but $\varepsilon_z$ only from 3.22 to 2.95.
\Cref{fig:ablation_camera_mechanism} illustrates this split on one image row.
The gap in lateral error grows with the distance to the principal point, where a wrong focal length moves a point the most.
During training, $\text{Rel}^\text{p}$ with intrinsics stalls near 5.0 without the encoding but keeps dropping to 3.94 with it.

\begin{figure}[t]
\centering
\includegraphics[width=\linewidth]{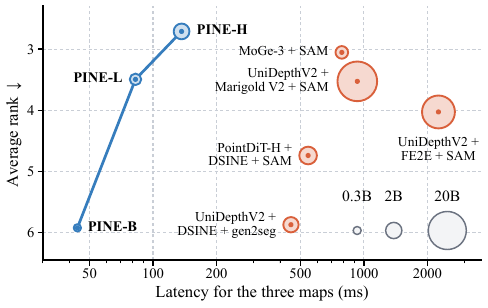}
\caption{\textbf{Average rank against latency for the three maps.}
Ranks are averaged over the columns of \cref{tab:per_dataset_point,tab:normal_comparison,tab:instance_comparison}, with each PINE model ranked against the baselines only.
Latency is measured on one A100 at $512 \times 512$ with batch size one.}
\label{fig:rank}
\end{figure}

\begin{table}[t]
\centering
\caption{\textbf{Region mass with and without grounding attention.}
Region mass in percent, averaged over all blocks and all query patches of a benchmark.
Chance is the region mass of uniform attention.
The blue number is the gain, the relative increase of the region mass above chance.}
\footnotesize
\newcommand{\gain}[1]{$_{\textcolor{cvprblue}{\uparrow#1\%}}$}
\setlength{\tabcolsep}{2pt}
\begin{tabular*}{\linewidth}{@{\extracolsep{\fill}}lcllll@{}}
\toprule
& & \multicolumn{2}{c}{Instance} & \multicolumn{2}{c}{Normal} \\
\cmidrule{3-4} \cmidrule{5-6}
Model & \shortstack[c]{Grounding\\attention} & \cocoexc & DRAM & NYUv2 & ScanNet \\
\midrule
\multicolumn{2}{@{}l}{Chance} & 24.5 & 33.7 & \phantom{0}5.1 & 12.9 \\
\midrule
\multirow{2}{*}{PINE-B} & \xmark & 35.5 & 44.3 & 11.7 & 19.8 \\
& \cmark & 36.6\gain{9.6} & 44.8\gain{4.5} & 13.2\gain{22.4} & 21.2\gain{21.0} \\
\midrule
\multirow{2}{*}{PINE-L} & \xmark & 32.8 & 41.4 & 10.1 & 18.1 \\
& \cmark & 33.9\gain{13.8} & 42.2\gain{11.0} & 11.1\gain{20.3} & 19.2\gain{22.0} \\
\midrule
\multirow{2}{*}{PINE-H} & \xmark & 33.5 & 43.9 & 10.8 & 18.9 \\
& \cmark & 34.5\gain{11.6} & 44.8\gain{8.8} & 11.5\gain{13.5} & 19.6\gain{11.3} \\
\bottomrule
\end{tabular*}
\label{tab:region_mass}
\end{table}

\paragraph{Stream Mask.}
In \cref{tab:ablation_components}, letting the instance tokens also attend to the geometry tokens raises $\text{Rel}^\text{p}$ without intrinsics from 4.82 to 4.93 and lowers the \cocoexc mIoU from 33.0 to 32.4.
PINE therefore isolates the instance stream.
Removing the normal-to-point attention raises the mean angular error with intrinsics from 17.81 to 17.94.
The normal stream thus keeps its attention to the point stream.

\paragraph{Grounding Attention.}
In \cref{tab:ablation_components}, removing grounding attention lowers the \cocoexc mIoU from 33.0 to 32.1 and raises the mean angular error from 18.23 to 18.47.
Injecting its code through adaLN instead lowers the DRAM mIoU from 54.8 to 52.3.
The region mass of a query patch is the share of its attention that falls on its own region.
\Cref{tab:region_mass} shows that grounding attention raises the region mass on all four benchmarks for all three models.
For PINE-L, the gain is 13.8\% on \cocoexc and 20.3\% on NYUv2.
\Cref{sec:supp_grounding} defines the regions and reports all nine benchmarks.

\subsection{Runtime Analysis}
\label{sec:runtime}
PINE-H has the best average rank over the three maps in \cref{fig:rank} and runs 5.7 times faster than MoGe-3 combined with SAM.
PINE-L matches the rank of UniDepthV2, Marigold V2 and SAM combined and runs 11 times faster.
PINE-H runs 3.1 times faster than SAM alone, while it also predicts the point map and the normals.
Marigold V2 with SAM needs 22.3B parameters and 1399\,ms, while PINE-H needs 1.85B and 137\,ms.
This shows the benefit of one forward pass over one model per map.

\section{Conclusion}
\label{sec:conclusion}

We have presented PINE, a one-step Omni Diffusion Transformer that predicts the point map, the normals and the instance map of an image in one forward pass.
It achieves state-of-the-art point maps and the best average rank on normals.
It exceeds SAM on large objects but still trails it on medium and small objects.
Camera-aware ray encoding exploits intrinsics when given, and grounding attention groups pixels by object.
Unlike chains of specialist models, its one-pass design suits interactive applications such as robotics.
Its point maps lack metric scale.
Future work can add metric scale and semantic classes as new streams.

\section*{LLM Usage Statement}
We used ChatGPT (OpenAI) to edit and restructure the text, format the \LaTeX{} source, tables and figure layout, and check citations and the consistency of reported numbers, and to refine code written by the authors.
The research ideas, method, experiments and results were conceived, implemented and verified by the authors, and the tools only edited or refined author-produced content.
The authors reviewed every suggestion and take full responsibility for the content of this paper.

{
    \small
    \bibliographystyle{ieeenat_fullname}
    \bibliography{main}
}

\input{X_suppl}

\end{document}

%% file: preamble.tex
\usepackage{xspace}
\newcommand{\cocoexc}{COCO$_\text{exc}$\xspace}
\definecolor{betterblue}{RGB}{0,90,181}
\definecolor{worseorange}{RGB}{220,110,0}

\usepackage{pifont}
\usepackage{algorithm}
\usepackage{algorithmic}
\newcommand{\cmark}{\ding{51}}
\newcommand{\xmark}{\ding{55}}

%% file: X_suppl.tex
\clearpage
\setcounter{page}{1}
\maketitlesupplementary
\appendix
\raggedbottom
\makeatletter
\setlength{\@fptop}{0pt}
\setlength{\@dblfptop}{0pt}
\makeatother
\renewcommand{\dbltopfraction}{0.9}
\renewcommand{\topfraction}{0.95}
\renewcommand{\textfraction}{0.05}
\setcounter{topnumber}{4}
\setcounter{dbltopnumber}{3}

\noindent This supplementary material extends the main paper with implementation and evaluation details, further analyses of the camera-aware ray encoding and grounding attention, complete quantitative results and qualitative comparisons.
It is organized as follows.
\begin{enumerate}
\item Implementation details (\cref{sec:supp_impl}) \dotfill \pageref{sec:supp_impl}
\item Evaluation details (\cref{sec:supp_eval}) \dotfill \pageref{sec:supp_eval}
\item Analysis of the camera-aware ray encoding (\cref{sec:supp_camera}) \dotfill \pageref{sec:supp_camera}
\item Analysis of grounding attention (\cref{sec:supp_grounding}) \dotfill \pageref{sec:supp_grounding}
\item Additional quantitative results (\cref{sec:supp_additional}) \dotfill \pageref{sec:supp_additional}
\item Qualitative results (\cref{sec:supp_qual}) \dotfill \pageref{sec:supp_qual}
\end{enumerate}

\section{Implementation Details}
\label{sec:supp_impl}

This section complements \cref{sec:objective,sec:exp_setup} of the main paper with the full training and inference setup.
\Cref{tab:supp_data} lists the training data.
The geometry streams train on the four synthetic datasets, and the instance stream also trains on the instance labels of EntitySeg and ADE20K.
On Hypersim, the instance loss uses only the scenes with at least 10 annotated objects, as in gen2seg~\cite{gen2seg}, which leaves 87K images with instance labels.
We remove the 555 EntitySeg images that come from COCO val2017, so that no test image of \cocoexc enters training.
\Cref{tab:supp_models,tab:supp_train,tab:supp_infer} list the model sizes, the training hyperparameters and the inference settings.

\subsection{Training Losses}
\label{sec:supp_loss}
The loss weights in \cref{eq:loss} are $\lambda_\text{rel} = 0.1$, $\lambda_\text{grad} = 0.5$ and $\lambda_\text{inst} = 0.02$.
The point-map target $x_\text{p}$ is normalized for shift and scale as in PointDiT~\cite{pointdit}.
In \cref{eq:loss_v}, the sky pixels add the same error, averaged separately and weighted by $\lambda_\text{sky} = 0.01$.
The gradient loss $\mathcal{L}_\text{grad}$ is the multi-scale gradient matching loss of MiDaS~\cite{midas} on the log-depth error, which sharpens depth edges.
The normal loss $\mathcal{L}_\text{sink}$ follows Marigold V2~\cite{marigoldv2}: inside every $5 \times 5$ window, a Sinkhorn plan matches the predicted normals to the target normals before the $L_1$ error is taken.
The instance loss $\mathcal{L}_\text{inst}$ follows gen2seg~\cite{gen2seg}, which adapts the discriminative loss of~\cite{instanceloss}.
It pulls the predicted colors of every instance to their mean, pushes all other pixels away from that mean, pushes the means of different instances apart and pulls the background to black.
Given that it uses the instance ids only to group pixels, the loss ignores which color an instance takes.

\begin{table}[t]
\centering
\caption{\textbf{Training data.} P, N and I mark point map, normal and instance labels.}
\small
\setlength{\tabcolsep}{4pt}
\begin{tabular*}{\columnwidth}{@{\extracolsep{\fill}}llrc@{}}
\toprule
Dataset & Domain & Images & Labels \\
\midrule
Hypersim~\cite{hypersim} & synthetic indoor & 39{,}241 & P, N, I \\
Virtual KITTI~2~\cite{vkitti2} & synthetic driving & 20{,}148 & P, N, I \\
UrbanSyn~\cite{urbansyn} & synthetic urban & 7{,}539 & P, N, I \\
Structured3D~\cite{structured3d} & synthetic indoor & 72{,}058 & P, N \\
EntitySeg~\cite{entityseg} & real & 6{,}369 & I \\
ADE20K~\cite{ade20k1,ade20k2} & real & 17{,}277 & I \\
\bottomrule
\end{tabular*}
\label{tab:supp_data}
\end{table}

\begin{table}[t]
\centering
\caption{\textbf{Model sizes.}}
\small
\setlength{\tabcolsep}{4pt}
\begin{tabular*}{\columnwidth}{@{\extracolsep{\fill}}lcccc@{}}
\toprule
Model & DINOv3 & DiT blocks & Width & Params (B) \\
\midrule
PINE-B & ViT-B/16 & 12 & 768 & 0.24 \\
PINE-L & ViT-L/16 & 24 & 1024 & 0.80 \\
PINE-H & ViT-H+/16 & 32 & 1280 & 1.85 \\
\bottomrule
\end{tabular*}
\label{tab:supp_models}
\end{table}

\begin{table}[t]
\centering
\caption{\textbf{Training hyperparameters.}}
\footnotesize
\setlength{\tabcolsep}{4pt}
\begin{tabular*}{\columnwidth}{@{\extracolsep{\fill}}p{0.43\columnwidth}p{0.53\columnwidth}@{}}
\toprule
Hyperparameter & Value \\
\midrule
Hardware & 8 $\times$ NVIDIA A100 \\
Training time & $\sim$20 hours for PINE-L \\
Initialization & PointDiT~\cite{pointdit} \\
Optimizer & AdamW, $\beta_1 = 0.9$, $\beta_2 = 0.95$ \\
Learning rate & $10^{-5}$, constant \\
Weight decay & 0 \\
Gradient clipping & 1.0, max $\ell_2$ norm \\
Epochs & 20 \\
Samples per epoch & 162{,}632 \\
Batch size & 64 \\
Precision & bfloat16 \\
EMA decay & 0.999 \\
\midrule
Noise scale $\sigma$ & 2 \\
Timestep distribution & $\operatorname{sigmoid}(\xi)$, $\xi \sim \mathcal{N}(-0.8, 0.8^2)$ \\
Probability of $t_s = 0$ & 0.1 \\
Clamp of $1 - t_s$ & 0.05 \\
$\lambda_\text{rel}$, $\lambda_\text{grad}$, $\lambda_\text{inst}$, $\lambda_\text{sky}$ & 0.1, 0.5, 0.02, 0.01 \\
Sinkhorn window & $5 \times 5$ \\
\midrule
Probability of keeping intrinsics & 0.8 \\
Reference focal length $f_\text{ref}$ & 420 pixels \\
Aspect ratio & log-uniform in $[0.5, 3.5]$ \\
Crop side & 0.6 to 1.0 of the largest crop \\
Resize & height to 512, probability 0.5, always below 512 \\
Horizontal flip & probability 0.5 \\
Color jitter & brightness, contrast, saturation 0.1, hue 0.05 \\
Blur, JPEG, autocontrast, equalization, channel permutation, grayscale & probability 0.2 each \\
\bottomrule
\end{tabular*}
\label{tab:supp_train}
\end{table}

\begin{table}[t]
\centering
\caption{\textbf{Inference settings.}}
\footnotesize
\setlength{\tabcolsep}{6pt}
\begin{tabular*}{\columnwidth}{@{\extracolsep{\fill}}ll@{}}
\toprule
Setting & Value \\
\midrule
Sampling steps & 1 \\
Timestep of all streams & 0 \\
Start of the geometry streams & $\mathcal{N}(\mathbf{0}, \sigma^2\mathbf{I})$ \\
Start of the instance stream & 0 \\
Without intrinsics & default camera with $f_\text{ref}$ \\
Focal recovery & as in MoGe~\cite{moge} \\
Weights & EMA \\
Precision & bfloat16 \\
\bottomrule
\end{tabular*}
\label{tab:supp_infer}
\end{table}

\subsection{Timesteps and Intrinsics}
The noise scale is $\sigma = 2$, and each geometry stream draws its own timestep $t_s = \operatorname{sigmoid}(\xi)$ with $\xi \sim \mathcal{N}(-0.8, 0.8^2)$.
With probability 0.1, $t_s$ is set to 0, the timestep of one-step inference.
Each training image keeps its intrinsics with probability 0.8 and otherwise uses the default camera of \cref{sec:ray_encoding}.

\section{Evaluation Details}
\label{sec:supp_eval}

This section details the evaluation protocol of \cref{sec:exp_setup}: the alignment, the metrics and the settings of the baselines.

\subsection{Point Maps and Depth}
We center-crop each image to $512 \times 512$.
We align each predicted point map $\hat{P}$ to the ground truth $P$ by a scale $s$ and a 3D shift $\mathbf{t}$.
For depth, we align the $z$ channel $\hat{z}$ of the predicted point map to the ground-truth depth $z$ by a separate scale $a$ and shift $b$.
We solve both alignments by the robust fit of MoGe~\cite{moge},
\begin{gather}
\min_{s, \mathbf{t}} \sum_{\mathbf{u} \in \Omega'} \frac{\|s \hat{P}(\mathbf{u}) + \mathbf{t} - P(\mathbf{u})\|_1}{\|P(\mathbf{u})\|}, \\
\min_{a, b} \sum_{\mathbf{u} \in \Omega'} \frac{|a \hat{z}(\mathbf{u}) + b - z(\mathbf{u})|}{z(\mathbf{u})},
\end{gather}
where $\Omega'$ holds the valid pixels of a $64 \times 64$ nearest-neighbor downsampling.
On the valid pixels $\Omega$ at full resolution, the point map metrics are
\begin{gather}
\text{Rel}^\text{p} = \Big\langle \frac{\|\hat{P}(\mathbf{u}) - P(\mathbf{u})\|}{\|P(\mathbf{u})\|} \Big\rangle_\Omega, \\
\delta_1^\text{p} = \Big\langle \big[\, \|\hat{P}(\mathbf{u}) - P(\mathbf{u})\| < 0.25 \min\big(\|P(\mathbf{u})\|, \|\hat{P}(\mathbf{u})\|\big) \big] \Big\rangle_\Omega,
\end{gather}
where $\hat{P}$ is the aligned prediction and $[\cdot]$ is 1 if the condition holds and 0 otherwise.
$\text{Rel}^\text{d}$ and $\delta_1^\text{d}$ are the same two metrics with the aligned depth in place of the point map.

\subsection{Baselines}
We compare the point maps with PointDiT~\cite{pointdit}, MoGe~\cite{moge}, MoGe-2~\cite{moge2}, MoGe-3~\cite{moge3}, FoundationGeo~\cite{foundationgeo}, UniDepthV2~\cite{unidepthv2}, Depth Pro~\cite{depthpro}, DA3~\cite{da3}, GeometryCrafter~\cite{geometrycrafter} and PPD~\cite{ppd}.
We compare the normals with DSINE~\cite{dsine}, MoGe-3~\cite{moge3}, Marigold~\cite{marigold}, Marigold v1.1~\cite{marigold}, Marigold V2~\cite{marigoldv2}, GeoWizard~\cite{geowizard}, StableNormal~\cite{stablenormal}, GenPercept~\cite{genpercept}, Lotus-D~\cite{lotus}, Lotus-G~\cite{lotus}, E2E-FT~\cite{e2eft}, FE2E~\cite{fe2e}, TransNormal~\cite{transnormal} and TransNormal-2~\cite{transnormal2}.
We compare the instances with SAM~\cite{sam}, SimpleClick~\cite{simpleclick} and gen2seg~\cite{gen2seg}.
We compare all methods without intrinsics and, for the baselines that accept them, also with intrinsics.
We re-evaluate SAM and gen2seg with their released weights under our instance protocol and take the other instance results from gen2seg.
With intrinsics, MoGe-2, MoGe-3 and FoundationGeo receive the ground-truth horizontal field of view, because they accept no principal point.
UniDepthV2 and DA3 receive the ground-truth intrinsics.
Depth Pro back-projects its depth with the ground-truth intrinsics, because its focal length input only rescales the depth.
For depth, we align MoGe, MoGe-2 and MoGe-3 in their affine-invariant space and Marigold V2 on log depth.

\subsection{Field of View and Planes}
We recover the horizontal field of view from each predicted point map by the camera evaluation of MoGe-2, with each image at its benchmark resolution and aspect ratio.
For planarity, we align each prediction to the ground-truth depth by a least-squares scale and shift, on log depth for Marigold V2, before the official per-plane median scaling of iBims-1.
The 155 planes contain 115 walls, 15 tables and 25 floors, and the ground-truth depth itself gives 0.19\,cm and $0.20^\circ$.

\subsection{Normals}
The accuracy within $11.25^\circ$ is the fraction of valid pixels with an angular error below $11.25^\circ$.

\subsection{Instances}
The point prompt of each ground-truth instance is its interior point farthest from the instance boundary.
All methods use the same point prompts.
For PINE and gen2seg, we select a mask from the instance map by the rule of gen2seg~\cite{gen2seg}.
This rule takes the query color $\mathbf{q}$ as a Gaussian-weighted mean around the point prompt, with $\sigma = 0.01$.
It scores every pixel of color $\mathbf{c}$ by $1 / \|\mathbf{c} - \mathbf{q}\|$ and rescales the scores to $[0, 255]$.
A joint bilateral filter of diameter 9, with both $\sigma$ at 75, smooths the scores, and the pixels with a score above 3 form the mask.
SAM~\cite{sam} takes the point prompt as one positive point, and we keep the output mask with the highest predicted IoU.
We split \cocoexc into small, medium and large instances by the COCO area thresholds of $32^2$ and $96^2$ pixels.

\section{Camera-Aware Ray Encoding}
\label{sec:supp_camera}
This section extends the camera-aware ray encoding ablation of \cref{sec:ablation}.
\Cref{fig:ablation_camera_bins,fig:ablation_curves,fig:ablation_camera} analyze the camera-aware ray encoding ablation of \cref{tab:ablation_components} with intrinsics.

\subsection{Lateral and Depth Error}
Without intrinsics, both models place the tokens on a regular grid, but the camera-aware ray encoding still reduces $\text{Rel}^\text{p}$ from 5.05 to 4.82.
To locate the gain with intrinsics, we split the aligned error $\mathbf{e}$ of each valid pixel into a lateral part $\varepsilon_{xy} = \|(e_x, e_y)\| / \|P(\mathbf{u})\|$ and a depth part $\varepsilon_z = |e_z| / \|P(\mathbf{u})\|$.
Averaged over the valid pixels with intrinsics, the encoding reduces $\varepsilon_{xy}$ from 2.28 to 1.27, by 44\%, but $\varepsilon_z$ only from 3.22 to 2.95, by 8\%.
It thus mainly corrects the lateral position of each point, as \cref{fig:ablation_camera_mechanism} illustrates.
This matches the back-projection of \cref{sec:ray_encoding}: a wrong focal length $\hat{f}$ keeps the depth and scales $x$ and $y$ by $f / \hat{f}$.

\subsection{Error by Distance and Depth}
\begin{figure}[tp]
\centering
\includegraphics[width=\linewidth]{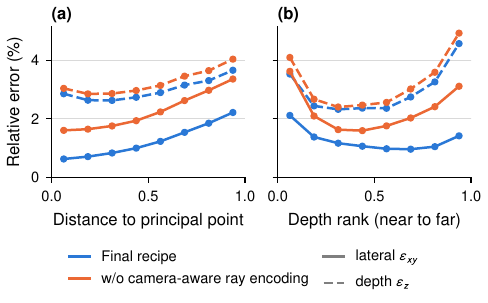}
\caption{\textbf{Error by distance and depth.}
Lateral error $\varepsilon_{xy}$ and depth error $\varepsilon_z$ in percent of $\|P(\mathbf{u})\|$, over the valid pixels of the $3{,}444$ images with intrinsics.
(a) Pixels in 8 bins of the distance to the principal point; (b) pixels in 8 bins of the depth rank.}
\label{fig:ablation_camera_bins}
\end{figure}

\begin{figure}[tp]
\centering
\includegraphics[width=\linewidth]{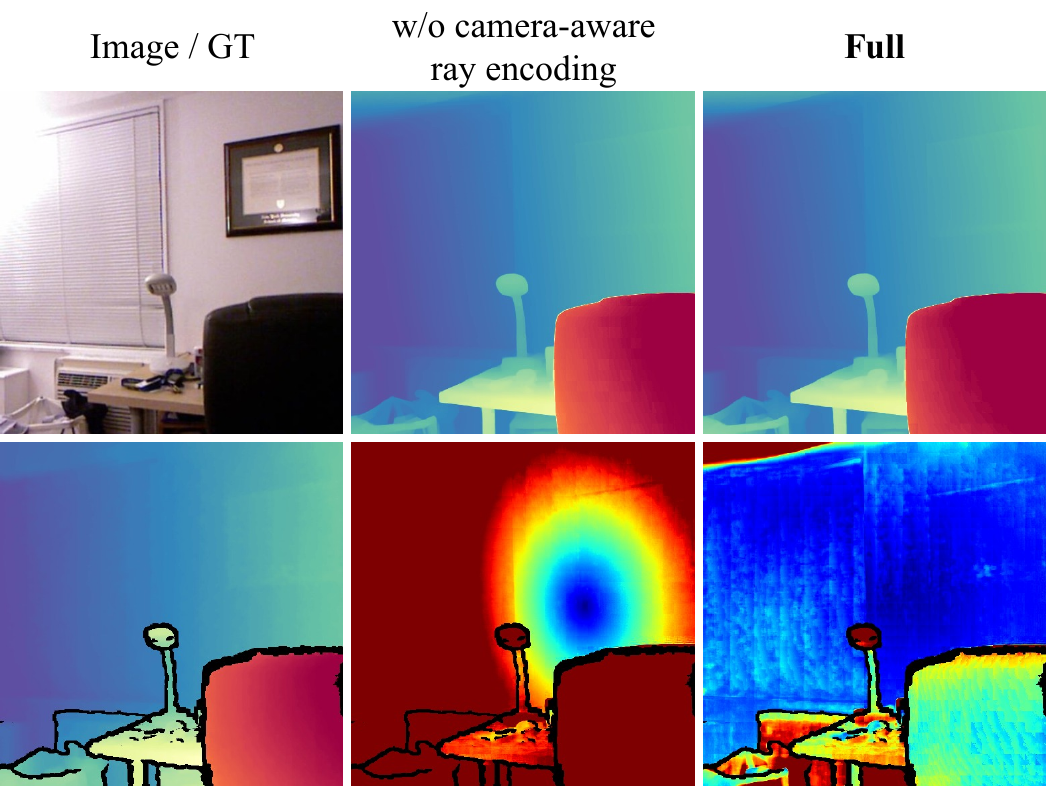}
\caption{\textbf{Point error with and without the camera-aware ray encoding.}
Top: input image and predicted depth. Bottom: ground-truth depth and point error maps.}
\label{fig:ablation_camera}
\end{figure}

\Cref{fig:ablation_camera_bins} bins the pixels by their distance to the principal point and by their depth rank.
The camera-aware ray encoding lowers the lateral error in every bin, while the depth error barely changes.
The gap in lateral error grows with the distance to the principal point.
This matches the back-projection of \cref{sec:ray_encoding}: a wrong focal length $\hat{f}$ scales $x$ and $y$ by $f / \hat{f}$, which moves a point more the farther it lies from the principal point.
\Cref{fig:ablation_camera} shows this pattern on one image.

\subsection{Training}
\begin{figure}[tp]
\centering
\includegraphics[width=\linewidth]{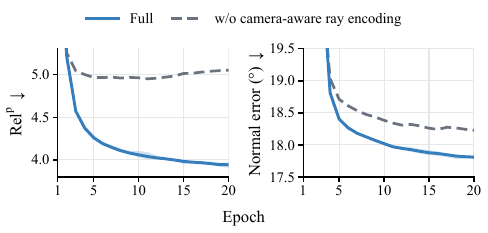}
\caption{\textbf{Effect of the camera-aware ray encoding during training.} $\text{Rel}^\text{p}$ and mean angular error with intrinsics over the training epochs of the ablation model in \cref{tab:ablation_components}, with and without the camera-aware ray encoding.}
\label{fig:ablation_curves}
\end{figure}

\Cref{fig:ablation_curves} shows that the gap is not a matter of training time.
Without the encoding, $\text{Rel}^\text{p}$ stalls near 5.0 after a few epochs, the value it also reaches without intrinsics.
With the encoding, $\text{Rel}^\text{p}$ keeps dropping to 3.94, and the mean angular error stays lower over all epochs.

\begin{table*}[tp]
\centering
\caption{\textbf{Region mass with and without grounding attention.}
Region mass in percent, averaged over all blocks and all query patches of a benchmark, and All pools the query patches of a stream.
Chance is the region mass of uniform attention.
The blue number is the gain, the relative increase of the region mass above chance.}
\footnotesize
\setlength{\tabcolsep}{2.5pt}
\newcommand{\gain}[1]{$_{\textcolor{cvprblue}{\uparrow#1\%}}$}
\resizebox{\textwidth}{!}{%
\begin{tabular}{ll*{11}{r@{}l}}
\toprule
& & \multicolumn{12}{c}{Instance} & \multicolumn{10}{c}{Normal} \\
\cmidrule(lr){3-14} \cmidrule(lr){15-24}
Model & \shortstack[l]{Grounding\\attention} & \multicolumn{2}{c}{\cocoexc} & \multicolumn{2}{c}{DRAM} & \multicolumn{2}{c}{EgoHOS} & \multicolumn{2}{c}{iShape} & \multicolumn{2}{c}{PIDRay} & \multicolumn{2}{c}{All} & \multicolumn{2}{c}{NYUv2} & \multicolumn{2}{c}{ScanNet} & \multicolumn{2}{c}{iBims-1} & \multicolumn{2}{c}{Sintel} & \multicolumn{2}{c}{All} \\
\midrule
\multicolumn{2}{l}{Chance} & 24.5 & & 33.7 & & 12.4 & & 8.6 & & 5.5 & & 17.0 & & 5.1 & & 12.9 & & 14.5 & & 3.6 & & 5.3 & \\
\midrule
\multirow{2}{*}{PINE-B} & w/o & 35.5 & & 44.3 & & 23.7 & & 13.7 & & 20.4 & & 27.0 & & 11.7 & & 19.8 & & 22.5 & & 6.5 & & 9.5 & \\
& with & 36.6 & \gain{9.6} & 44.8 & \gain{4.5} & 25.4 & \gain{14.9} & 14.4 & \gain{13.2} & 22.8 & \gain{15.5} & 28.1 & \gain{11.7} & 13.2 & \gain{22.4} & 21.2 & \gain{21.0} & 24.2 & \gain{21.9} & 7.5 & \gain{32.4} & 10.6 & \gain{26.8} \\
\midrule
\multirow{2}{*}{PINE-L} & w/o & 32.8 & & 41.4 & & 20.0 & & 12.0 & & 17.3 & & 24.3 & & 10.1 & & 18.1 & & 20.9 & & 5.8 & & 8.4 & \\
& with & 33.9 & \gain{13.8} & 42.2 & \gain{11.0} & 21.5 & \gain{19.9} & 12.7 & \gain{20.7} & 19.0 & \gain{15.2} & 25.4 & \gain{15.6} & 11.1 & \gain{20.3} & 19.2 & \gain{22.0} & 22.1 & \gain{19.4} & 6.5 & \gain{35.2} & 9.3 & \gain{27.3} \\
\midrule
\multirow{2}{*}{PINE-H} & w/o & 33.5 & & 43.9 & & 20.5 & & 12.1 & & 17.2 & & 24.8 & & 10.8 & & 18.9 & & 21.6 & & 6.2 & & 9.0 & \\
& with & 34.5 & \gain{11.6} & 44.8 & \gain{8.8} & 21.7 & \gain{15.3} & 12.5 & \gain{11.8} & 18.6 & \gain{11.8} & 25.8 & \gain{12.0} & 11.5 & \gain{13.5} & 19.6 & \gain{11.3} & 22.5 & \gain{12.6} & 6.7 & \gain{22.0} & 9.6 & \gain{17.1} \\
\bottomrule
\end{tabular}}
\label{tab:supp_region_mass}
\end{table*}

\section{Grounding Attention}
\label{sec:supp_grounding}
\begin{figure}[tp]
\centering
\includegraphics[width=\linewidth]{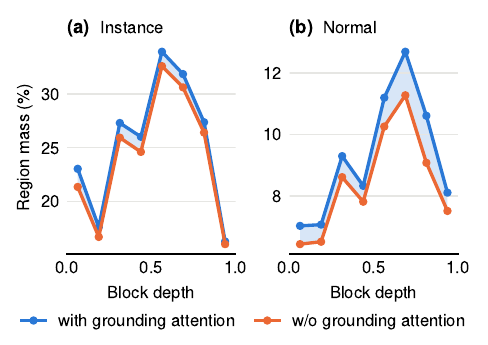}
\caption{\textbf{Region mass over the depth of PINE-L.}
Region mass with and without grounding attention for (a) the instance stream and (b) the normal stream.
Each point averages three consecutive blocks and pools the query patches of all benchmarks of the stream.}
\label{fig:supp_region_mass_blocks}
\end{figure}

This section extends the grounding attention ablation of \cref{sec:ablation}.
\Cref{tab:supp_region_mass,fig:supp_region_mass_blocks,fig:supp_grounding_samples} analyze how the grounding attention of \cref{sec:grounding} changes the attention.

\subsection{Region Mass}
The region mass of a query patch is the share of its attention that falls on its own region.
For the instance stream, the region is the ground-truth instance that covers more than half of the patch.
For the normal stream, the region is the connected set of patches whose normals lie within $10^\circ$ of the normal of the query patch.
We run each model once per image and compute the attention of every block twice, with and without grounding attention, that is, with and without the bias in the logits.
Chance is the region mass of uniform attention, the area share of the region.
The gain is the relative increase of the region mass above chance.
\Cref{tab:supp_region_mass} shows that grounding attention raises the region mass on all nine benchmarks for all three models.
For PINE-L, the gain is 15.6\% on the instance stream and 27.3\% on the normal stream.
This agrees with \cref{tab:ablation_components}, where removing grounding attention lowers the \cocoexc mIoU from 33.0 to 32.1 and raises the mean angular error from 18.23 to 18.47.

\subsection{Blocks}
\Cref{fig:supp_region_mass_blocks} plots the region mass of PINE-L over the depth of the network.
Grounding attention raises the region mass at every depth for both streams.
For the normal stream, the increase is largest in the second half of the network, at 1.5 percentage points.

\subsection{Attention Maps}
\begin{figure*}[tp]
\centering
\includegraphics[width=\textwidth]{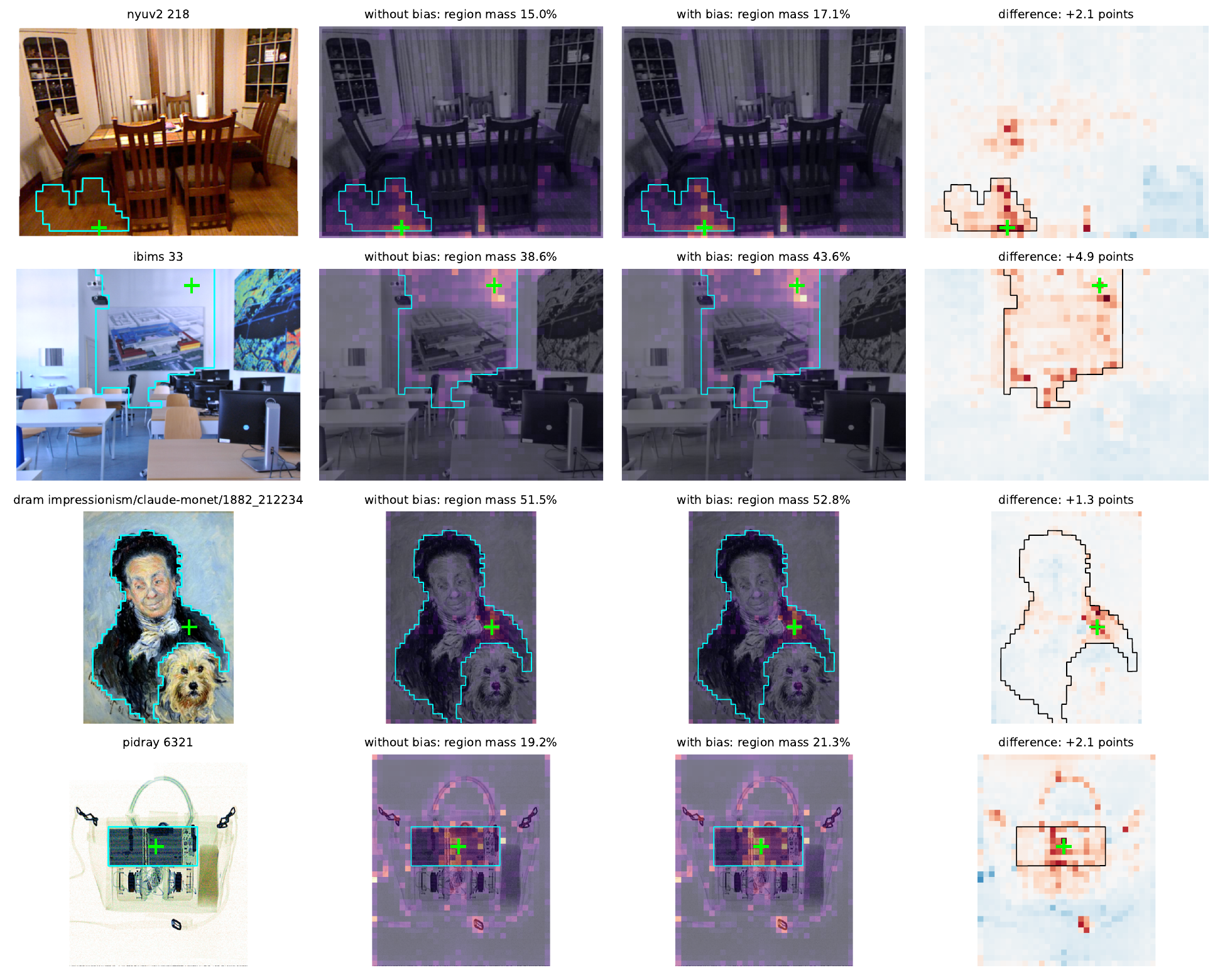}
\caption{\textbf{Attention with and without grounding attention.}
Each row shows the image, the attention of the query patch without and with grounding attention, and their difference.
The cross marks the query patch, and the contour marks its region.}
\label{fig:supp_grounding_samples}
\end{figure*}

\Cref{fig:supp_grounding_samples} shows the attention of one query patch of PINE-H on four selected images, averaged over all blocks.
For the normal stream, the query patch is the patch with the largest region.
For the instance stream, it is the middle patch of the largest instance.
As in \cref{fig:grounding}, grounding attention concentrates the attention of the query patch on its own region.

\section{Additional Results}
\label{sec:supp_additional}

This section gives the complete quantitative results behind the averages and summaries of \cref{sec:exp_results}.

\begin{table}[t]
\centering
\renewcommand{\arraystretch}{0.9}
\caption{\textbf{Field-of-view estimation.}
Mean absolute error in degrees of the horizontal field of view recovered from the predicted point map without intrinsics.
The Avg column is the sample-weighted mean over all $3{,}444$ images.}
\setlength{\tabcolsep}{4pt}
\resizebox{\columnwidth}{!}{%
\begin{tabular}{lcccccccc}
\toprule
Method & DIODE & KITTI & NYUv2 & ETH3D & HAMMER & iBims-1 & Booster & Avg \\
\midrule
MoGe-2~\cite{moge2} & \textbf{6.61} & 10.64 & \textbf{1.60} & \textbf{3.73} & 12.18 & 4.77 & 9.88 & 7.28 \\
MoGe-3~\cite{moge3} & 7.70 & 12.02 & 1.61 & 4.00 & 13.75 & 5.54 & 11.78 & 8.22 \\
FoundationGeo~\cite{foundationgeo} & 8.01 & \textbf{4.56} & 2.57 & 4.43 & 12.18 & 4.49 & 8.68 & 6.70 \\
\midrule
PointDiT-B~\cite{pointdit} & 8.11 & 17.65 & 5.54 & 20.11 & 17.48 & 6.70 & 8.12 & 13.08 \\
PointDiT-L~\cite{pointdit} & 8.34 & 14.02 & 4.48 & 11.00 & 13.16 & 5.35 & 8.48 & 10.03 \\
PointDiT-H~\cite{pointdit} & 9.38 & 19.70 & 6.88 & 12.92 & 19.30 & 6.81 & 7.26 & 13.46 \\
\midrule
PINE-B & 7.97 & 5.76 & 7.30 & 9.90 & 6.66 & 6.26 & 7.59 & 7.33 \\
PINE-L & 7.34 & 5.45 & 7.69 & 7.80 & 4.97 & 4.93 & \textbf{5.16} & 6.48 \\
PINE-H & 7.76 & 4.91 & 4.01 & 6.97 & \textbf{3.70} & \textbf{3.86} & 5.63 & \textbf{5.35} \\
\bottomrule
\end{tabular}}
\label{tab:fov}
\end{table}

\begin{table}[t]
\centering
\renewcommand{\arraystretch}{0.9}
\caption{\textbf{Cost of single models} for one $512 \times 512$ image with batch size one on one A100.
P, D, N and I mark point map, depth, normal and instance outputs, and SAM and SimpleClick take one point prompt.}
\footnotesize
\setlength{\tabcolsep}{3pt}
\begin{tabular*}{\columnwidth}{@{\extracolsep{\fill}}llcc@{}}
\toprule
Method & Output & Params (B) & Latency (ms) $\downarrow$ \\
\midrule
Depth Pro~\cite{depthpro} & P & 0.95 & 909.3 \\
UniDepthV2~\cite{unidepthv2} & P & 0.35 & 23.3 \\
DA3~\cite{da3} & P & 1.36 & 97.3 \\
MoGe-2~\cite{moge2} & P+N & 0.33 & 67.9 \\
MoGe-3~\cite{moge3} & P+N & 0.37 & 360.5 \\
FoundationGeo~\cite{foundationgeo} & P & 0.31 & 90.1 \\
MoGe~\cite{moge} & P & 0.31 & 86.0 \\
GeometryCrafter~\cite{geometrycrafter} & P & 2.67 & 964.7 \\
PPD (4 steps)~\cite{ppd} & P & 0.80 & 691.9 \\
\midrule
Lotus-G~\cite{lotus} & N & 1.29 & 202.7 \\
DSINE~\cite{dsine} & N & 0.07 & 56.1 \\
FE2E~\cite{fe2e} & D+N & 12.5 & 1808.3 \\
Marigold V2~\cite{marigoldv2} (depth) & D & 21.4 & 491.6 \\
Marigold V2~\cite{marigoldv2} (normal) & N & 21.4 & 482.0 \\
\midrule
SAM (ViT-H)~\cite{sam} & I & 0.64 & 425.1 \\
SimpleClick (ViT-H)~\cite{simpleclick} & I & 0.66 & 189.2 \\
gen2seg~\cite{gen2seg} & I & 1.29 & 371.5 \\
\midrule
PointDiT-B~\cite{pointdit} & P & 0.22 & 23.3 \\
PointDiT-L~\cite{pointdit} & P & 0.77 & 45.8 \\
PointDiT-H~\cite{pointdit} & P & 1.81 & 62.5 \\
\midrule
PINE-B & P+N+I & 0.24 & 43.8 \\
PINE-L & P+N+I & 0.80 & 82.6 \\
PINE-H & P+N+I & 1.85 & 136.6 \\
\bottomrule
\end{tabular*}
\label{tab:cost_single}
\end{table}

\begin{table}[t]
\centering
\renewcommand{\arraystretch}{0.9}
\caption{\textbf{Cost of the three maps.}
Each baseline pipeline runs one model per map, and its latency sums the latencies in \cref{tab:cost_single}.
Marigold V2 runs its depth and normal models, which share 21.3B parameters.}
\small
\setlength{\tabcolsep}{4pt}
\resizebox{\columnwidth}{!}{
\begin{tabular}{lcc}
\toprule
Pipeline & Params (B) & Latency (ms) $\downarrow$ \\
\midrule
MoGe-2~\cite{moge2} + SAM~\cite{sam} & 0.97 & 493 \\
PointDiT-H~\cite{pointdit} + DSINE~\cite{dsine} + SAM~\cite{sam} & 2.52 & 544 \\
MoGe-3~\cite{moge3} + SAM~\cite{sam} & 1.01 & 786 \\
UniDepthV2~\cite{unidepthv2} + Marigold V2~\cite{marigoldv2} (N) + SAM~\cite{sam} & 22.4 & 930 \\
Marigold V2~\cite{marigoldv2} (D+N) + SAM~\cite{sam} & 22.3 & 1399 \\
FE2E~\cite{fe2e} + SAM~\cite{sam} & 13.2 & 2233 \\
\midrule
PINE-H & 1.85 & \textbf{137} \\
\bottomrule
\end{tabular}
}
\label{tab:cost_combo}
\end{table}

\begin{table}[t]
\centering
\renewcommand{\arraystretch}{0.9}
\caption{\textbf{Sampling steps.}
Point-map and depth metrics over the 3{,}444 images and normal metrics averaged over the four normal benchmarks.
Best results within each model are in bold.}
\footnotesize
\setlength{\tabcolsep}{4pt}
\resizebox{\columnwidth}{!}{%
\begin{tabular}{llcccccccccc}
\toprule
\multirow{2}{*}{Model} & \multirow{2}{*}{Steps} & \multicolumn{2}{c}{Rel$^\text{p}$ $\downarrow$} & $\delta_1^\text{p}$ $\uparrow$ & \multicolumn{2}{c}{Rel$^\text{d}$ $\downarrow$} & $\delta_1^\text{d}$ $\uparrow$ & \multicolumn{2}{c}{Mean $\downarrow$} & \multicolumn{2}{c}{$11.25^\circ$ $\uparrow$} \\
\cmidrule(lr){3-4} \cmidrule(lr){5-5} \cmidrule(lr){6-7} \cmidrule(lr){8-8} \cmidrule(lr){9-10} \cmidrule(lr){11-12}
& & w/o $K$ & with $K$ & w/o $K$ & w/o $K$ & with $K$ & w/o $K$ & w/o $K$ & with $K$ & w/o $K$ & with $K$ \\
\midrule
\multirow{4}{*}{PINE-B} & 1 & \textbf{5.62} & 4.69 & \textbf{96.75} & 3.70 & 3.60 & \textbf{97.75} & \textbf{19.63} & \textbf{19.34} & 53.7 & \textbf{54.5} \\
& 2 & \textbf{5.62} & \textbf{4.65} & 96.74 & \textbf{3.69} & \textbf{3.58} & 97.72 & 19.84 & 19.55 & \textbf{53.8} & \textbf{54.5} \\
& 3 & 5.69 & 4.69 & 96.70 & 3.75 & 3.63 & 97.67 & 20.11 & 19.81 & 53.5 & 54.3 \\
& 4 & 5.77 & 4.75 & 96.65 & 3.82 & 3.68 & 97.61 & 20.33 & 20.02 & 53.3 & 54.0 \\
\midrule
\multirow{4}{*}{PINE-L} & 1 & \textbf{4.70} & 3.76 & \textbf{97.78} & \textbf{3.04} & \textbf{2.92} & \textbf{98.29} & \textbf{18.17} & \textbf{17.68} & \textbf{57.2} & \textbf{58.4} \\
& 2 & 4.72 & \textbf{3.73} & 97.74 & \textbf{3.04} & \textbf{2.92} & 98.27 & 18.40 & 17.91 & 57.0 & 58.2 \\
& 3 & 4.77 & 3.78 & 97.70 & 3.09 & 2.95 & 98.24 & 18.66 & 18.17 & 56.7 & 57.9 \\
& 4 & 4.83 & 3.82 & 97.65 & 3.14 & 3.00 & 98.20 & 18.86 & 18.37 & 56.5 & 57.7 \\
\midrule
\multirow{4}{*}{PINE-H} & 1 & 4.16 & \textbf{3.33} & \textbf{98.13} & \textbf{2.79} & \textbf{2.65} & \textbf{98.51} & \textbf{17.31} & \textbf{16.98} & \textbf{59.2} & \textbf{59.9} \\
& 2 & \textbf{4.13} & \textbf{3.33} & \textbf{98.13} & 2.80 & \textbf{2.65} & 98.49 & 17.53 & 17.20 & 58.9 & 59.7 \\
& 3 & 4.19 & 3.38 & 98.08 & 2.84 & 2.69 & 98.47 & 17.77 & 17.44 & 58.6 & 59.4 \\
& 4 & 4.24 & 3.42 & 98.05 & 2.88 & 2.73 & 98.41 & 17.95 & 17.62 & 58.4 & 59.1 \\
\bottomrule
\end{tabular}}
\label{tab:supp_steps}
\end{table}

\begin{table}[t]
\centering
\caption{\textbf{Planarity on iBims-1.}
Planarity error $\varepsilon^\text{plan}$ in cm and orientation error $\varepsilon^\text{orie}$ in degrees over 155 planes of 100 images.
All models run without intrinsics on the full frame.}
\footnotesize
\setlength{\tabcolsep}{2pt}
\resizebox{\columnwidth}{!}{%
\begin{tabular}{lcccccccc}
\toprule
 & \multicolumn{2}{c}{All} & \multicolumn{2}{c}{Wall} & \multicolumn{2}{c}{Table} & \multicolumn{2}{c}{Floor} \\
\cmidrule(lr){2-3} \cmidrule(lr){4-5} \cmidrule(lr){6-7} \cmidrule(lr){8-9}
Method & $\varepsilon^\text{plan}$$\downarrow$ & $\varepsilon^\text{orie}$$\downarrow$ & $\varepsilon^\text{plan}$$\downarrow$ & $\varepsilon^\text{orie}$$\downarrow$ & $\varepsilon^\text{plan}$$\downarrow$ & $\varepsilon^\text{orie}$$\downarrow$ & $\varepsilon^\text{plan}$$\downarrow$ & $\varepsilon^\text{orie}$$\downarrow$ \\
\midrule
MoGe-2~\cite{moge2} & 0.57 & \textbf{1.92} & 0.64 & \textbf{2.05} & \textbf{0.31} & 1.81 & 0.42 & \textbf{1.39} \\
MoGe-3~\cite{moge3} & \textbf{0.44} & 1.96 & \textbf{0.48} & \textbf{2.05} & \textbf{0.31} & 1.89 & \textbf{0.34} & 1.58 \\
FoundationGeo~\cite{foundationgeo} & 0.83 & 2.42 & 0.96 & 2.69 & 0.36 & \textbf{1.78} & 0.52 & 1.57 \\
Marigold V2~\cite{marigoldv2} & 0.78 & 2.50 & 0.87 & 2.77 & 0.47 & 2.22 & 0.58 & 1.44 \\
\midrule
PointDiT-B~\cite{pointdit} & 1.00 & 3.00 & 1.13 & 3.20 & 0.66 & 2.49 & 0.61 & 2.37 \\
PointDiT-L~\cite{pointdit} & 0.82 & 2.35 & 0.94 & 2.61 & 0.38 & 1.80 & 0.52 & 1.53 \\
PointDiT-H~\cite{pointdit} & 0.62 & 2.30 & 0.68 & 2.38 & 0.32 & 2.49 & 0.52 & 1.84 \\
\midrule
PINE-B & 0.85 & 2.53 & 0.87 & 2.47 & 0.77 & 3.09 & 0.79 & 2.46 \\
PINE-L & 0.61 & 2.10 & 0.63 & 2.20 & 0.35 & 2.18 & 0.66 & 1.63 \\
PINE-H & 0.54 & 2.10 & 0.56 & 2.14 & 0.37 & 2.13 & 0.54 & 1.89 \\
\bottomrule
\end{tabular}}
\label{tab:ibims_planarity}
\end{table}

\begin{table*}[tp]
\renewcommand{\arraystretch}{0.86}
\centering
\caption{\textbf{Per-dataset point map results.}
The Avg column is the sample-weighted mean over all $3{,}444$ images.}
\setlength{\tabcolsep}{3pt}
\resizebox{\textwidth}{!}{
\begin{tabular}{lcccccccccccccccc}
\toprule
\multirow{2}{*}{Method} & \multicolumn{2}{c}{DIODE} & \multicolumn{2}{c}{KITTI} & \multicolumn{2}{c}{NYUv2} & \multicolumn{2}{c}{ETH3D} & \multicolumn{2}{c}{HAMMER} & \multicolumn{2}{c}{iBims-1} & \multicolumn{2}{c}{Booster} & \multicolumn{2}{c}{Avg} \\
\cmidrule(lr){2-3} \cmidrule(lr){4-5} \cmidrule(lr){6-7} \cmidrule(lr){8-9} \cmidrule(lr){10-11} \cmidrule(lr){12-13} \cmidrule(lr){14-15} \cmidrule(lr){16-17}
 & Rel$^\text{p}$$\downarrow$ & $\delta_1^\text{p}$$\uparrow$ & Rel$^\text{p}$$\downarrow$ & $\delta_1^\text{p}$$\uparrow$ & Rel$^\text{p}$$\downarrow$ & $\delta_1^\text{p}$$\uparrow$ & Rel$^\text{p}$$\downarrow$ & $\delta_1^\text{p}$$\uparrow$ & Rel$^\text{p}$$\downarrow$ & $\delta_1^\text{p}$$\uparrow$ & Rel$^\text{p}$$\downarrow$ & $\delta_1^\text{p}$$\uparrow$ & Rel$^\text{p}$$\downarrow$ & $\delta_1^\text{p}$$\uparrow$ & Rel$^\text{p}$$\downarrow$ & $\delta_1^\text{p}$$\uparrow$ \\
\midrule
\rowcolor{gray!15}\multicolumn{17}{l}{\textbf{\textit{Without intrinsics}}} \\
GeometryCrafter~\cite{geometrycrafter} & 5.88 & 95.06 & 4.84 & 97.97 & 4.10 & 98.64 & 5.46 & 97.03 & 6.75 & 95.46 & 5.50 & 97.23 & 3.73 & 99.31 & 5.45 & 96.75 \\
PPD~\cite{ppd} & 6.35 & 93.68 & 7.04 & 96.16 & 4.10 & 98.27 & 5.08 & 97.36 & 5.19 & 97.69 & 4.55 & 97.97 & 3.32 & 98.73 & 5.54 & 96.59 \\
Depth Pro~\cite{depthpro} & 6.36 & 94.11 & 7.33 & 95.97 & 4.08 & 98.48 & 5.90 & 96.85 & 5.27 & 98.03 & 4.54 & 98.11 & 2.87 & 99.81 & 5.71 & 96.71 \\
UniDepthV2~\cite{unidepthv2} & 6.02 & 94.35 & \textbf{4.03} & \textbf{98.26} & 3.29 & 98.52 & 4.02 & 98.79 & 4.57 & 97.53 & 4.07 & 98.34 & 3.78 & 98.90 & 4.45 & 97.35 \\
DA3~\cite{da3} & 5.47 & 93.68 & 6.37 & 96.72 & 3.67 & 98.20 & 3.44 & 98.20 & 4.60 & 96.95 & 4.08 & 97.81 & 2.92 & 99.03 & 4.77 & 96.63 \\
MoGe~\cite{moge} & \textbf{4.25} & \textbf{96.37} & 4.10 & 98.10 & 3.48 & 98.63 & \textbf{3.35} & 98.99 & 5.51 & 95.95 & \textbf{3.95} & 97.82 & 2.74 & 99.06 & 4.21 & 97.45 \\
MoGe-2~\cite{moge2} & 4.57 & 95.28 & 5.87 & 97.86 & 3.13 & 98.83 & 3.73 & 98.71 & 5.19 & 97.19 & 4.00 & 98.27 & \textbf{2.59} & \textbf{99.90} & 4.53 & 97.46 \\
MoGe-3~\cite{moge3} & 4.91 & 95.14 & 6.04 & 98.11 & \textbf{3.09} & \textbf{98.84} & 3.67 & \textbf{99.02} & 5.91 & 96.53 & 4.05 & 97.91 & 2.98 & 99.38 & 4.79 & 97.36 \\
FoundationGeo~\cite{foundationgeo} & 5.44 & 95.07 & 5.21 & 98.01 & 3.43 & 98.80 & 3.83 & 98.67 & 4.71 & 98.41 & 4.08 & 98.61 & 3.08 & 99.61 & 4.57 & 97.71 \\
PointDiT-B~\cite{pointdit} & 6.75 & 93.63 & 5.54 & 96.93 & 5.08 & 97.49 & 6.46 & 96.84 & 5.68 & 98.49 & 4.96 & 97.75 & 3.72 & 99.74 & 5.84 & 96.69 \\
PointDiT-L~\cite{pointdit} & 5.88 & 94.81 & 5.25 & 97.44 & 4.32 & 98.09 & 5.02 & 97.53 & 4.23 & 99.17 & 4.24 & 98.10 & 3.11 & 99.88 & 4.90 & 97.42 \\
PointDiT-H~\cite{pointdit} & 5.27 & 95.87 & 5.21 & 97.79 & 3.93 & 98.37 & 4.53 & 97.93 & 3.51 & 99.55 & \textbf{3.95} & 98.52 & 3.06 & 99.86 & 4.45 & 97.93 \\
PINE-B & 6.60 & 93.17 & 6.54 & 96.64 & 5.21 & 97.45 & 4.77 & 97.86 & 4.86 & 98.95 & 5.26 & 97.19 & 3.60 & 99.69 & 5.62 & 96.75 \\
PINE-L & 5.39 & 95.53 & 5.40 & 97.36 & 4.87 & 97.91 & 4.22 & 98.59 & 3.74 & 99.60 & 4.20 & 98.40 & 2.75 & 99.88 & 4.70 & 97.78 \\
PINE-H & 5.00 & 96.16 & 5.23 & 97.74 & 4.10 & 98.21 & 3.46 & 98.97 & \textbf{2.96} & \textbf{99.70} & 4.11 & \textbf{98.75} & 2.72 & 99.89 & \textbf{4.16} & \textbf{98.13} \\
\midrule
\rowcolor{gray!15}\multicolumn{17}{l}{\textbf{\textit{With intrinsics}}} \\
Depth Pro~\cite{depthpro} & 5.64 & 95.07 & 4.30 & 97.49 & 3.93 & 98.48 & 5.09 & 97.22 & 4.00 & 98.36 & 3.59 & 98.33 & 2.71 & \textbf{99.92} & 4.53 & 97.35 \\
UniDepthV2~\cite{unidepthv2} & 4.77 & 95.81 & \textbf{3.32} & \textbf{98.36} & 3.14 & 98.80 & \textbf{3.11} & \textbf{99.02} & 3.84 & 98.27 & 3.28 & 98.54 & 3.11 & 98.82 & 3.70 & 97.95 \\
DA3~\cite{da3} & 4.93 & 94.16 & 4.47 & 97.40 & 3.49 & 98.13 & 3.21 & 98.26 & 3.58 & 98.01 & 3.05 & 98.08 & 2.93 & 99.00 & 3.96 & 97.10 \\
MoGe-2~\cite{moge2} & 4.25 & 96.00 & 5.56 & 97.91 & \textbf{3.05} & 98.86 & 3.47 & 98.82 & 5.30 & 96.85 & 3.84 & 98.63 & 2.65 & 99.58 & 4.38 & 97.58 \\
MoGe-3~\cite{moge3} & 4.68 & 95.41 & 6.36 & 97.96 & \textbf{3.05} & 98.85 & 3.51 & 98.95 & 5.59 & 96.40 & 3.96 & 97.98 & 2.81 & 99.37 & 4.70 & 97.35 \\
FoundationGeo~\cite{foundationgeo} & 5.19 & 95.53 & 5.27 & 98.06 & 3.35 & \textbf{98.87} & 3.67 & 98.77 & 4.31 & 98.40 & 4.00 & 98.57 & 3.00 & 99.59 & 4.40 & 97.85 \\
PINE-B & 5.70 & 93.88 & 4.88 & 97.16 & 4.43 & 97.68 & 4.70 & 97.94 & 3.85 & 98.94 & 4.41 & 97.31 & 2.99 & 99.74 & 4.69 & 97.06 \\
PINE-L & 4.52 & 95.59 & 4.39 & 97.54 & 3.46 & 98.26 & 4.11 & 98.63 & 2.68 & 99.61 & 3.06 & 98.55 & 2.19 & 99.77 & 3.76 & 97.90 \\
PINE-H & \textbf{4.14} & \textbf{96.13} & 4.04 & 97.84 & 3.18 & 98.53 & 3.43 & 98.98 & \textbf{2.11} & \textbf{99.73} & \textbf{2.99} & \textbf{98.80} & \textbf{2.11} & 99.89 & \textbf{3.33} & \textbf{98.21} \\
\bottomrule
\end{tabular}
}
\label{tab:per_dataset_point}
\vspace{0.4em}
\centering
\caption{\textbf{Quantitative comparison on zero-shot surface normal estimation with all baselines.}
Mean angular error in degrees and accuracy within $11.25^\circ$ under the full-frame protocol of DSINE.
$\star$: relies on pretrained Stable Diffusion; $\ddagger$: the Marigold normal model of its repository.}
\setlength{\tabcolsep}{3pt}
\resizebox{\textwidth}{!}{
\begin{tabular}{lccccccccc}
\toprule
Method &
\multicolumn{2}{c}{NYUv2 (Indoor)} &
\multicolumn{2}{c}{ScanNet (Indoor)} &
\multicolumn{2}{c}{iBims-1 (Indoor)} &
\multicolumn{2}{c}{Sintel (Outdoor)} &
{\begin{tabular}[c]{@{}c@{}}Avg.\\Rank $\downarrow$\end{tabular}}
\\

\cmidrule(lr){2-3}
\cmidrule(lr){4-5}
\cmidrule(lr){6-7}
\cmidrule(lr){8-9}

&
MeanErr $\downarrow$ &
$11.25^\circ$ $\uparrow$ &
MeanErr $\downarrow$ &
$11.25^\circ$ $\uparrow$ &
MeanErr $\downarrow$ &
$11.25^\circ$ $\uparrow$ &
MeanErr $\downarrow$ &
$11.25^\circ$ $\uparrow$ &
\\

\midrule
\rowcolor{gray!15}\multicolumn{10}{l}{\textbf{\textit{Without intrinsics}}} \\
Marigold~\cite{marigold}$^{\ddagger\star}$ & 20.9 & 50.5 & 21.3 & 45.6 & 18.5 & 64.7 & -- & -- & 16.5 \\
GeoWizard~\cite{geowizard}$^{\star}$ & 18.9 & 50.7 & 17.4 & 53.8 & 19.3 & 63.0 & 40.3 & 12.3 & 15.9 \\
GenPercept~\cite{genpercept}$^{\star}$ & 18.2 & 56.3 & 17.7 & 58.3 & 18.2 & 64.0 & 37.6 & 16.2 & 14.4 \\
StableNormal~\cite{stablenormal}$^{\star}$ & 18.6 & 53.5 & 17.1 & 57.4 & 18.2 & 65.0 & 36.7 & 14.1 & 14.3 \\
TransNormal~\cite{transnormal}$^{\star}$ & 16.6 & 59.4 & 15.5 & 62.4 & 17.2 & 67.2 & 35.1 & 18.7 & 12.0 \\
Lotus-G~\cite{lotus} & 16.5 & 59.4 & 15.1 & 63.9 & 17.2 & 66.2 & 33.6 & 21.0 & 11.2 \\
DSINE~\cite{dsine} & 16.4 & 59.6 & 16.2 & 61.0 & 17.1 & 67.4 & 34.9 & 21.5 & 10.8 \\
Lotus-D~\cite{lotus}$^{\star}$ & 16.2 & 59.8 & 14.7 & 64.0 & 17.1 & 66.4 & 32.3 & 22.4 & 8.8 \\
E2E-FT~\cite{e2eft}$^{\star}$ & 16.5 & 60.4 & 14.7 & 66.1 & 16.1 & 69.7 & 33.5 & 22.3 & 8.3 \\
Marigold v1.1~\cite{marigold} & 16.1 & 60.5 & 14.5 & 66.1 & 16.3 & 68.5 & -- & -- & 7.2 \\
FE2E~\cite{fe2e} & 16.2 & 59.6 & 13.8 & 67.2 & 15.1 & 70.6 & 31.2 & 22.3 & 6.1 \\
Marigold V2~\cite{marigoldv2} & 16.0 & 61.6 & 13.8 & 67.9 & 15.2 & \textbf{71.9} & 29.7 & 27.3 & 4.3 \\
TransNormal-2~\cite{transnormal2}$^{\star}$ & 14.9 & 62.4 & 12.9 & 69.1 & 15.1 & 70.1 & 28.2 & 28.1 & 3.1 \\
MoGe-3~\cite{moge3} & \textbf{14.7} & 62.3 & \textbf{12.8} & 68.9 & \textbf{14.8} & 70.1 & 26.4 & 29.1 & 2.4 \\
PINE-B & 16.4 & 59.2 & 15.3 & 63.0 & 16.5 & 68.5 & 30.4 & 24.1 & 9.3 \\
PINE-L & 15.7 & 61.9 & 14.1 & 66.4 & 15.5 & 70.9 & 27.5 & 29.6 & 4.3 \\
PINE-H & 15.3 & \textbf{63.1} & 13.3 & \textbf{69.2} & 15.1 & 71.8 & \textbf{25.6} & \textbf{32.5} & \textbf{1.9} \\
\midrule
\rowcolor{gray!15}\multicolumn{10}{l}{\textbf{\textit{With intrinsics}}} \\
PINE-B & 16.3 & 59.7 & 14.9 & 64.7 & 16.2 & 69.0 & 30.0 & 24.6 & -- \\
PINE-L & 15.6 & 62.5 & 13.6 & 68.5 & 15.1 & 71.7 & 26.5 & 30.9 & -- \\
PINE-H & \textbf{15.1} & \textbf{63.5} & \textbf{13.0} & \textbf{70.0} & \textbf{14.8} & \textbf{72.2} & \textbf{25.0} & \textbf{34.0} & -- \\
\bottomrule
\end{tabular}
}
\label{tab:normal_comparison_full}
\end{table*}

\begin{table*}[tp]
\renewcommand{\arraystretch}{0.9}
\centering
\caption{\textbf{Per-dataset depth results.}
The Avg column is the sample-weighted mean over all $3{,}444$ images.}
\setlength{\tabcolsep}{3pt}
\resizebox{\textwidth}{!}{
\begin{tabular}{lcccccccccccccccc}
\toprule
\multirow{2}{*}{Method} & \multicolumn{2}{c}{DIODE} & \multicolumn{2}{c}{KITTI} & \multicolumn{2}{c}{NYUv2} & \multicolumn{2}{c}{ETH3D} & \multicolumn{2}{c}{HAMMER} & \multicolumn{2}{c}{iBims-1} & \multicolumn{2}{c}{Booster} & \multicolumn{2}{c}{Avg} \\
\cmidrule(lr){2-3} \cmidrule(lr){4-5} \cmidrule(lr){6-7} \cmidrule(lr){8-9} \cmidrule(lr){10-11} \cmidrule(lr){12-13} \cmidrule(lr){14-15} \cmidrule(lr){16-17}
 & Rel$^\text{d}$$\downarrow$ & $\delta_1^\text{d}$$\uparrow$ & Rel$^\text{d}$$\downarrow$ & $\delta_1^\text{d}$$\uparrow$ & Rel$^\text{d}$$\downarrow$ & $\delta_1^\text{d}$$\uparrow$ & Rel$^\text{d}$$\downarrow$ & $\delta_1^\text{d}$$\uparrow$ & Rel$^\text{d}$$\downarrow$ & $\delta_1^\text{d}$$\uparrow$ & Rel$^\text{d}$$\downarrow$ & $\delta_1^\text{d}$$\uparrow$ & Rel$^\text{d}$$\downarrow$ & $\delta_1^\text{d}$$\uparrow$ & Rel$^\text{d}$$\downarrow$ & $\delta_1^\text{d}$$\uparrow$ \\
\midrule
\rowcolor{gray!15}\multicolumn{17}{l}{\textbf{\textit{Without intrinsics}}} \\
GeometryCrafter~\cite{geometrycrafter} & 3.74 & 96.89 & 3.73 & 98.20 & 2.99 & 98.82 & 3.26 & 98.38 & 3.86 & 97.24 & 2.95 & 98.06 & 2.25 & 99.14 & 3.52 & 97.84 \\
PPD~\cite{ppd} & 4.69 & 95.86 & 4.81 & 97.47 & 3.57 & 98.38 & 4.15 & 97.69 & 2.57 & 99.33 & 3.11 & 98.44 & 2.56 & 99.21 & 3.88 & 97.78 \\
Depth Pro~\cite{depthpro} & 4.40 & 95.84 & 4.10 & 97.59 & 3.34 & 98.65 & 4.16 & 97.65 & 2.98 & 98.83 & 2.83 & 98.60 & 1.88 & \textbf{99.92} & 3.72 & 97.74 \\
UniDepthV2~\cite{unidepthv2} & 3.62 & 97.11 & 3.25 & 98.44 & 2.60 & 98.84 & 2.65 & 99.00 & 2.23 & 99.42 & 2.46 & 98.49 & 2.27 & 98.98 & 2.86 & 98.52 \\
DA3~\cite{da3} & 3.73 & 96.26 & 3.96 & 97.40 & 2.96 & 98.46 & 2.74 & 98.45 & 2.74 & 98.63 & 2.54 & 98.35 & 2.22 & 99.42 & 3.22 & 97.81 \\
MoGe~\cite{moge} & 3.17 & 97.32 & 3.26 & 98.33 & 2.64 & 98.90 & 2.53 & 99.17 & 3.75 & 96.92 & 2.57 & 98.14 & 2.21 & 99.16 & 3.10 & 98.01 \\
MoGe-2~\cite{moge2} & \textbf{2.89} & 97.79 & 3.10 & 98.48 & \textbf{2.54} & 98.95 & 2.58 & 99.12 & 3.24 & 98.38 & \textbf{2.17} & 98.91 & 1.94 & 99.67 & 2.87 & 98.50 \\
MoGe-3~\cite{moge3} & 3.00 & 97.62 & \textbf{3.04} & \textbf{98.51} & 2.56 & 98.95 & 2.49 & 99.28 & 3.35 & 98.10 & 2.18 & \textbf{98.94} & 1.99 & 99.50 & 2.90 & 98.43 \\
FoundationGeo~\cite{foundationgeo} & 3.38 & 97.43 & 3.32 & 98.44 & 2.69 & \textbf{99.03} & 2.72 & \textbf{99.37} & 2.67 & 98.66 & 2.49 & 98.86 & 2.08 & 99.61 & 2.95 & 98.52 \\
Marigold V2~\cite{marigoldv2} & 4.48 & 97.55 & 3.96 & 98.31 & 3.50 & 98.82 & 3.46 & 98.98 & 2.52 & 99.34 & 2.89 & 98.84 & 1.97 & 99.64 & 3.55 & \textbf{98.59} \\
PointDiT-B~\cite{pointdit} & 4.08 & 96.75 & 4.28 & 97.28 & 3.58 & 98.01 & 3.43 & 97.92 & 3.23 & 99.01 & 3.02 & 98.52 & 2.63 & 99.72 & 3.70 & 97.84 \\
PointDiT-L~\cite{pointdit} & 3.48 & 97.29 & 4.07 & 97.58 & 3.05 & 98.47 & 2.96 & 98.19 & 2.40 & 99.33 & 2.49 & 98.73 & 1.99 & 99.87 & 3.15 & 98.22 \\
PointDiT-H~\cite{pointdit} & 3.01 & \textbf{97.82} & 3.91 & 97.88 & 2.79 & 98.65 & 2.64 & 98.45 & 1.90 & 99.54 & 2.38 & 98.73 & 1.84 & 99.86 & 2.81 & 98.51 \\
PINE-B & 4.28 & 96.12 & 4.61 & 97.17 & 3.61 & 98.00 & 3.38 & 98.20 & 2.75 & 99.20 & 3.19 & 98.29 & 2.43 & 99.74 & 3.70 & 97.75 \\
PINE-L & 3.40 & 97.02 & 4.19 & 97.45 & 3.00 & 98.48 & 2.71 & 98.89 & 2.07 & 99.62 & 2.49 & 98.67 & 1.89 & 99.88 & 3.04 & 98.29 \\
PINE-H & 3.17 & 97.42 & 4.00 & 97.85 & 2.80 & 98.57 & \textbf{2.41} & 99.06 & \textbf{1.71} & \textbf{99.67} & 2.35 & 98.83 & \textbf{1.80} & 99.88 & \textbf{2.79} & 98.51 \\
\midrule
\rowcolor{gray!15}\multicolumn{17}{l}{\textbf{\textit{With intrinsics}}} \\
Depth Pro~\cite{depthpro} & 4.39 & 95.84 & 4.10 & 97.59 & 3.34 & 98.65 & 4.18 & 97.65 & 2.98 & 98.83 & 2.83 & 98.60 & 1.88 & \textbf{99.92} & 3.72 & 97.74 \\
UniDepthV2~\cite{unidepthv2} & 3.70 & 96.94 & 3.20 & 98.39 & 2.61 & 98.87 & 2.64 & 99.16 & 2.18 & 99.48 & 2.63 & 98.61 & 2.23 & 99.06 & 2.87 & 98.52 \\
DA3~\cite{da3} & 3.73 & 96.20 & 3.89 & 97.46 & 3.01 & 98.38 & 2.73 & 98.47 & 2.52 & 99.08 & 2.57 & 98.28 & 2.31 & 99.27 & 3.17 & 97.89 \\
MoGe-2~\cite{moge2} & \textbf{2.89} & \textbf{97.79} & 3.10 & 98.48 & \textbf{2.54} & 98.95 & 2.58 & 99.12 & 3.24 & 98.37 & \textbf{2.17} & 98.91 & 1.94 & 99.66 & 2.87 & 98.50 \\
MoGe-3~\cite{moge3} & 3.00 & 97.62 & \textbf{3.04} & \textbf{98.51} & 2.56 & 98.95 & 2.49 & 99.29 & 3.35 & 98.10 & 2.18 & \textbf{98.94} & 1.99 & 99.50 & 2.90 & 98.43 \\
FoundationGeo~\cite{foundationgeo} & 3.38 & 97.43 & 3.32 & 98.44 & 2.69 & \textbf{99.03} & 2.72 & \textbf{99.37} & 2.67 & 98.66 & 2.49 & 98.86 & 2.08 & 99.61 & 2.95 & 98.52 \\
PINE-B & 4.21 & 95.99 & 4.39 & 97.23 & 3.57 & 98.03 & 3.39 & 98.19 & 2.62 & 99.28 & 3.18 & 98.17 & 2.29 & 99.83 & 3.60 & 97.75 \\
PINE-L & 3.37 & 97.02 & 4.04 & 97.60 & 2.90 & 98.50 & 2.71 & 98.87 & 1.77 & 99.63 & 2.45 & 98.55 & 1.73 & 99.76 & 2.92 & 98.32 \\
PINE-H & 3.10 & 97.39 & 3.78 & 97.92 & 2.72 & 98.64 & \textbf{2.41} & 99.05 & \textbf{1.41} & \textbf{99.72} & 2.35 & 98.66 & \textbf{1.59} & 99.89 & \textbf{2.65} & \textbf{98.54} \\
\bottomrule
\end{tabular}
}
\label{tab:per_dataset_depth}
\end{table*}

\subsection{Point Maps}
\label{sec:supp_per_dataset}
\Cref{tab:per_dataset_point} gives the per-dataset point map results that \cref{tab:comparison} averages.
PINE-H has the lowest average $\text{Rel}^\text{p}$ in both settings without leading every dataset.
Without intrinsics, it has the lowest $\text{Rel}^\text{p}$ on HAMMER, ranks second on Booster and third on ETH3D.
With intrinsics, it has the lowest $\text{Rel}^\text{p}$ on DIODE, HAMMER, iBims-1 and Booster.

\subsection{Surface Normals}
\Cref{tab:normal_comparison_full} adds Marigold, Marigold v1.1, GenPercept, TransNormal and Lotus-G to \cref{tab:normal_comparison}.
We evaluate MoGe-3, Marigold V2, TransNormal, TransNormal-2 and PINE ourselves and take the other results from FE2E~\cite{fe2e}.
The average rank is the mean rank over the eight metrics among the methods without intrinsics.
PINE-H keeps the best average rank among all these methods.

\subsection{Depth}
\Cref{tab:per_dataset_depth} gives the per-dataset depth results that \cref{tab:comparison} averages.
Without intrinsics, PINE-H has the lowest $\text{Rel}^\text{d}$ on ETH3D, HAMMER and Booster.

\subsection{Field of View}
\Cref{tab:fov} gives the per-dataset field-of-view errors behind \cref{tab:fov_main}.
At every model size, PINE lowers the average error of PointDiT, from $13.08^\circ$ to $7.33^\circ$ for B, from $10.03^\circ$ to $6.48^\circ$ for L and from $13.46^\circ$ to $5.35^\circ$ for H.

\subsection{Planarity}
\Cref{tab:ibims_planarity} gives the planarity on iBims-1.
At every model size, PINE lowers the planarity error of PointDiT over all planes, from 1.00 to 0.85\,cm for B, from 0.82 to 0.61\,cm for L and from 0.62 to 0.54\,cm for H.
MoGe-3 keeps the lowest planarity error, 0.44\,cm.

\subsection{Cost}
\Cref{tab:cost_single,tab:cost_combo} give the parameters and latencies behind \cref{fig:rank}.
PINE-H predicts all three maps in 137\,ms, while the fastest baseline pipeline, MoGe-2 with SAM, takes 493\,ms.
One forward pass thus costs less than one model per map.

\subsection{Sampling Steps}
\Cref{tab:supp_steps} evaluates PINE with one to four sampling steps.
From one to two steps, $\text{Rel}^\text{p}$ changes by at most 0.04 for every model, and more steps raise all errors slightly.
The mean angular error rises with every added step.
PINE therefore uses one sampling step.

\section{Qualitative Results}
\label{sec:supp_qual}

\begin{figure*}[tp]
\centering
\includegraphics[width=\textwidth]{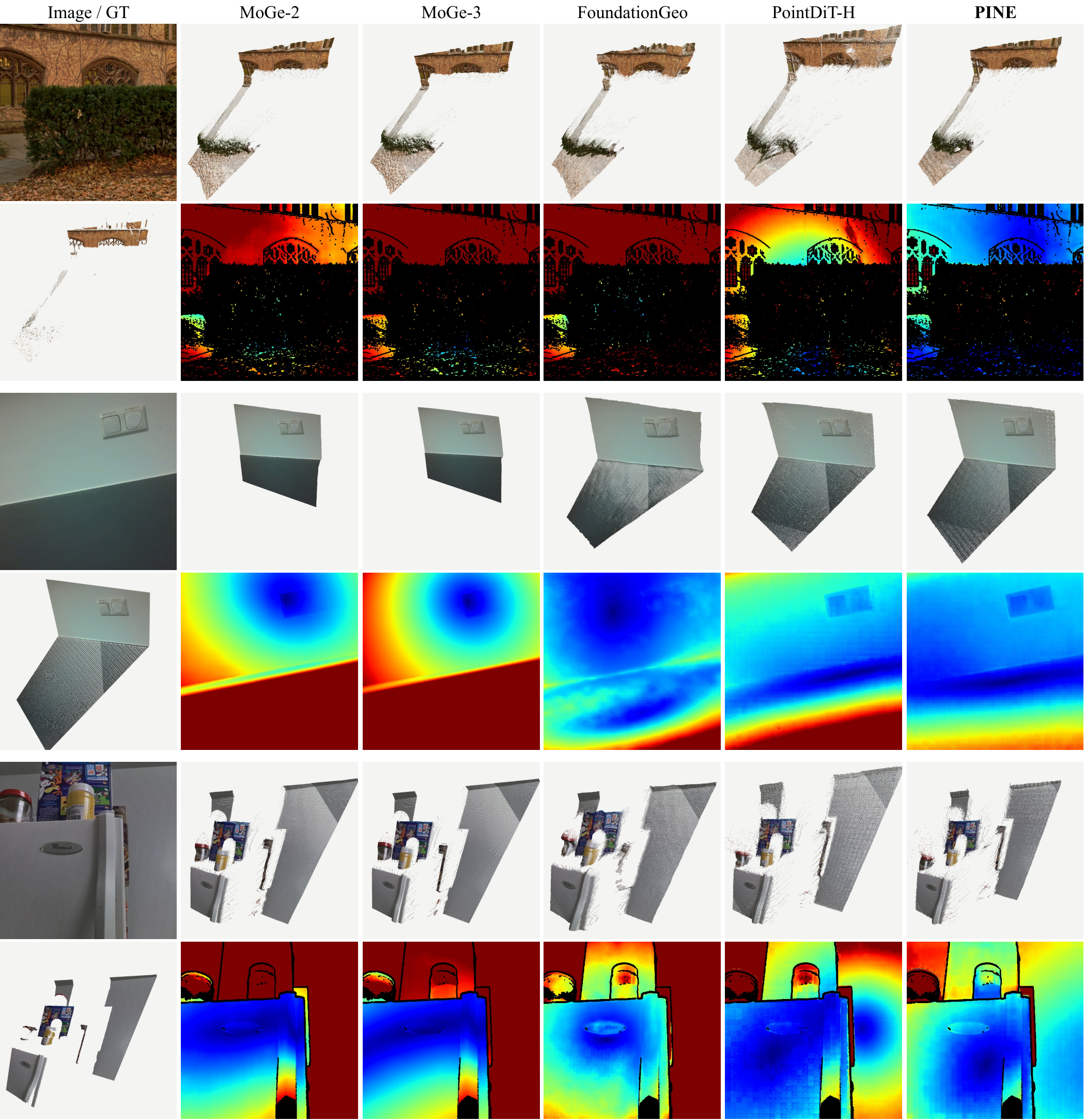}
\caption{\textbf{Point maps (1/3).} For each image, the first row shows the input and the predicted point maps, and the second row shows the ground-truth point map and the point error map of each method.}
\label{fig:supp_point_err_1}
\end{figure*}

\begin{figure*}[tp]
\centering
\includegraphics[width=\textwidth]{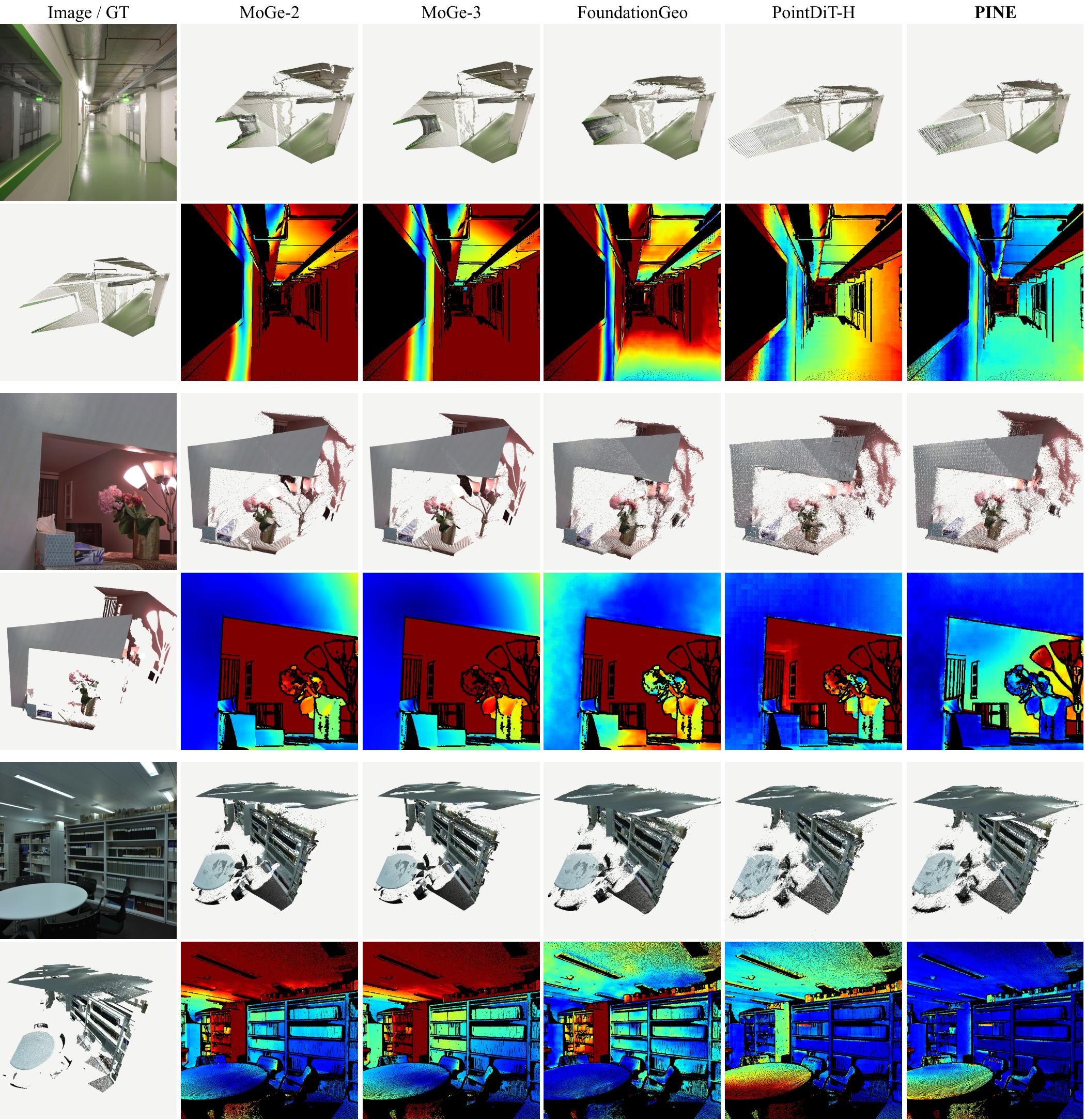}
\caption{\textbf{Point maps (2/3).} For each image, the first row shows the input and the predicted point maps, and the second row shows the ground-truth point map and the point error map of each method.}
\label{fig:supp_point_err_2}
\end{figure*}

\begin{figure*}[tp]
\centering
\includegraphics[width=\textwidth]{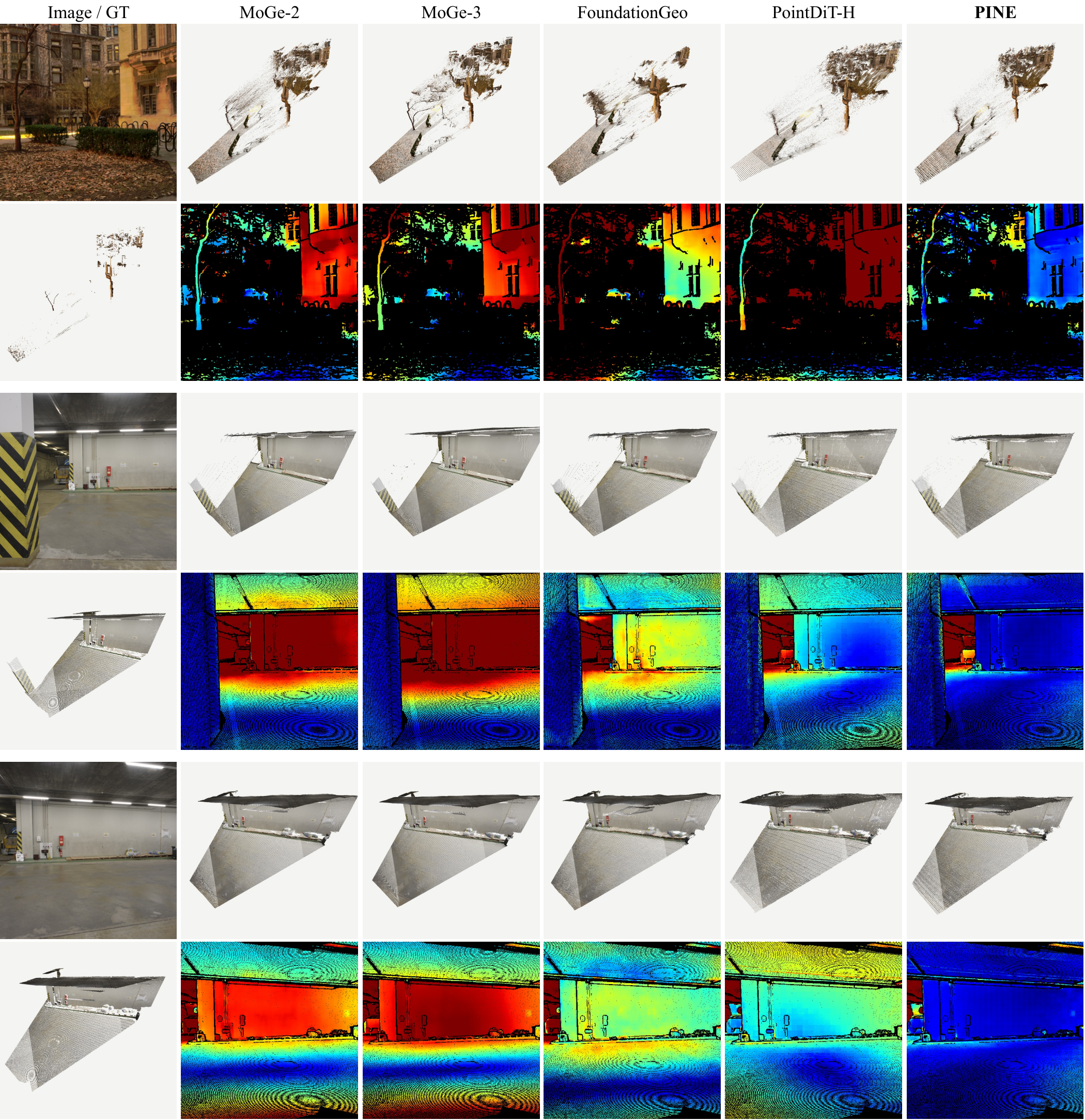}
\caption{\textbf{Point maps (3/3).} For each image, the first row shows the input and the predicted point maps, and the second row shows the ground-truth point map and the point error map of each method.}
\label{fig:supp_point_err_3}
\end{figure*}

\begin{figure*}[tp]
\centering
\includegraphics[width=\textwidth]{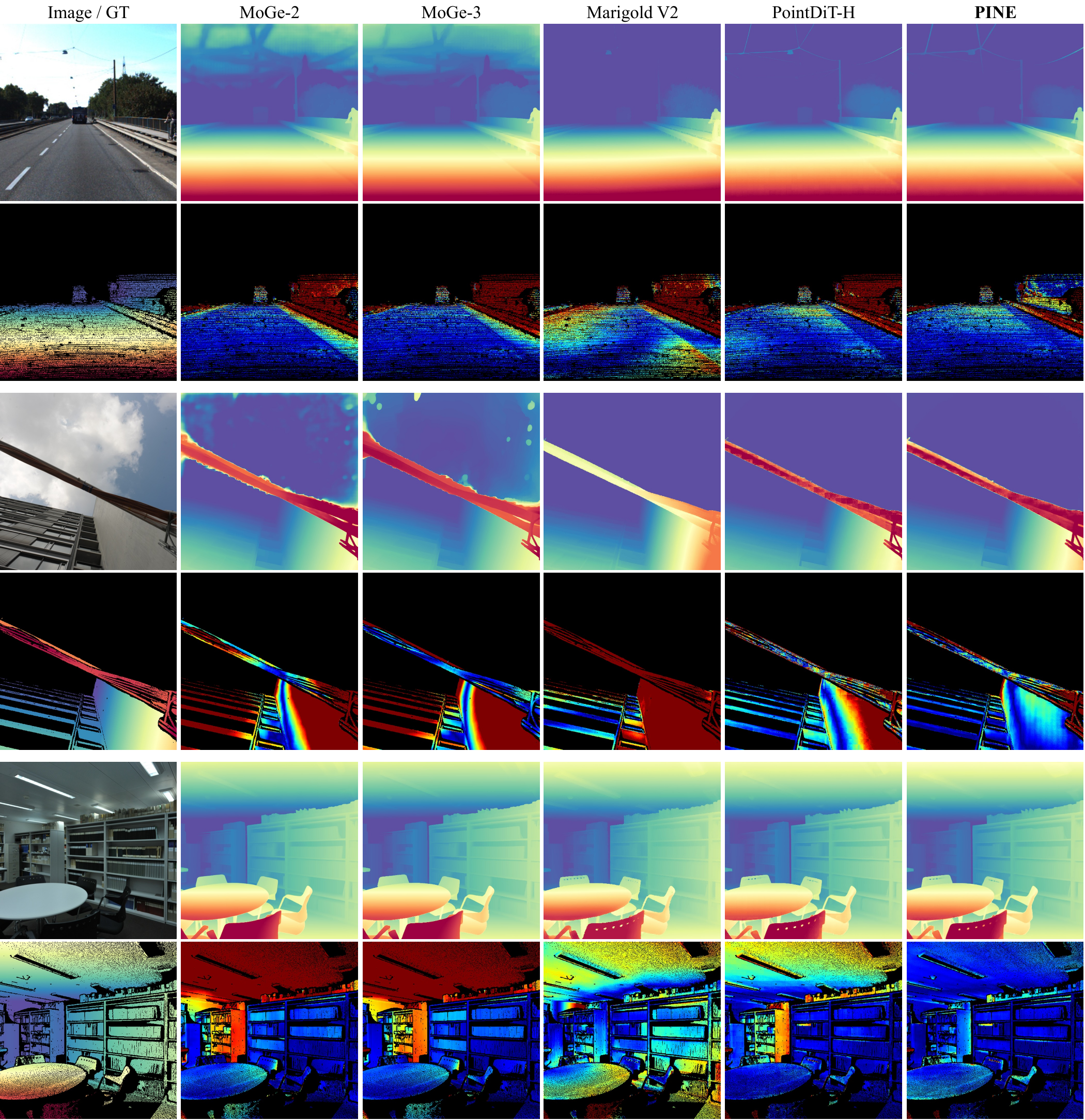}
\caption{\textbf{Depth.} For each image, the first row shows the input and the predicted depth maps, and the second row shows the ground-truth depth and the depth error map of each method.}
\label{fig:supp_depth_err}
\end{figure*}

\begin{figure*}[tp]
\centering
\includegraphics[width=\textwidth]{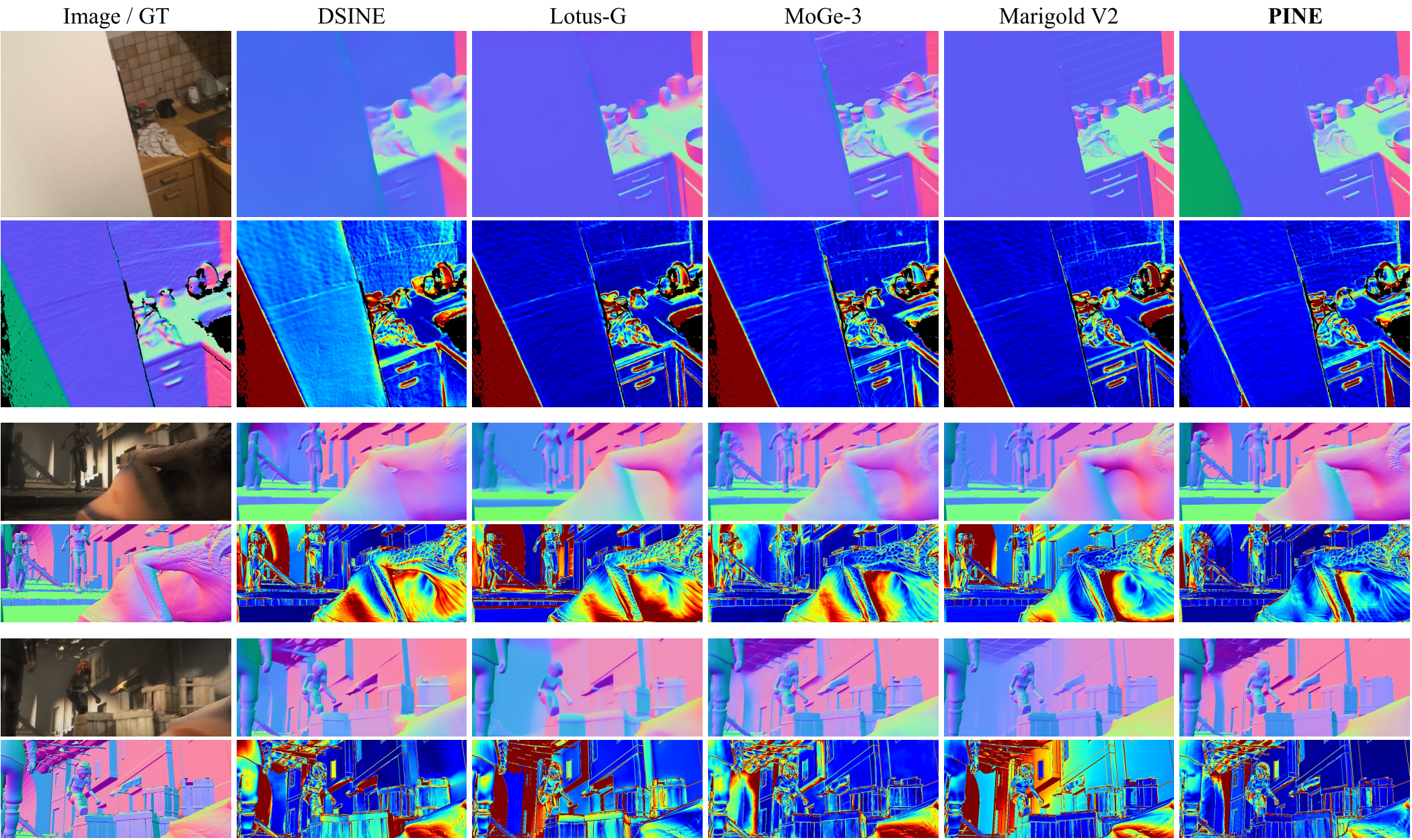}
\caption{\textbf{Surface normals.} For each image, the first row shows the input and the predicted normals, and the second row shows the ground-truth normals and the angular error map of each method.}
\label{fig:supp_normal_err}
\end{figure*}

\begin{figure*}[tp]
\centering
\includegraphics[width=\textwidth]{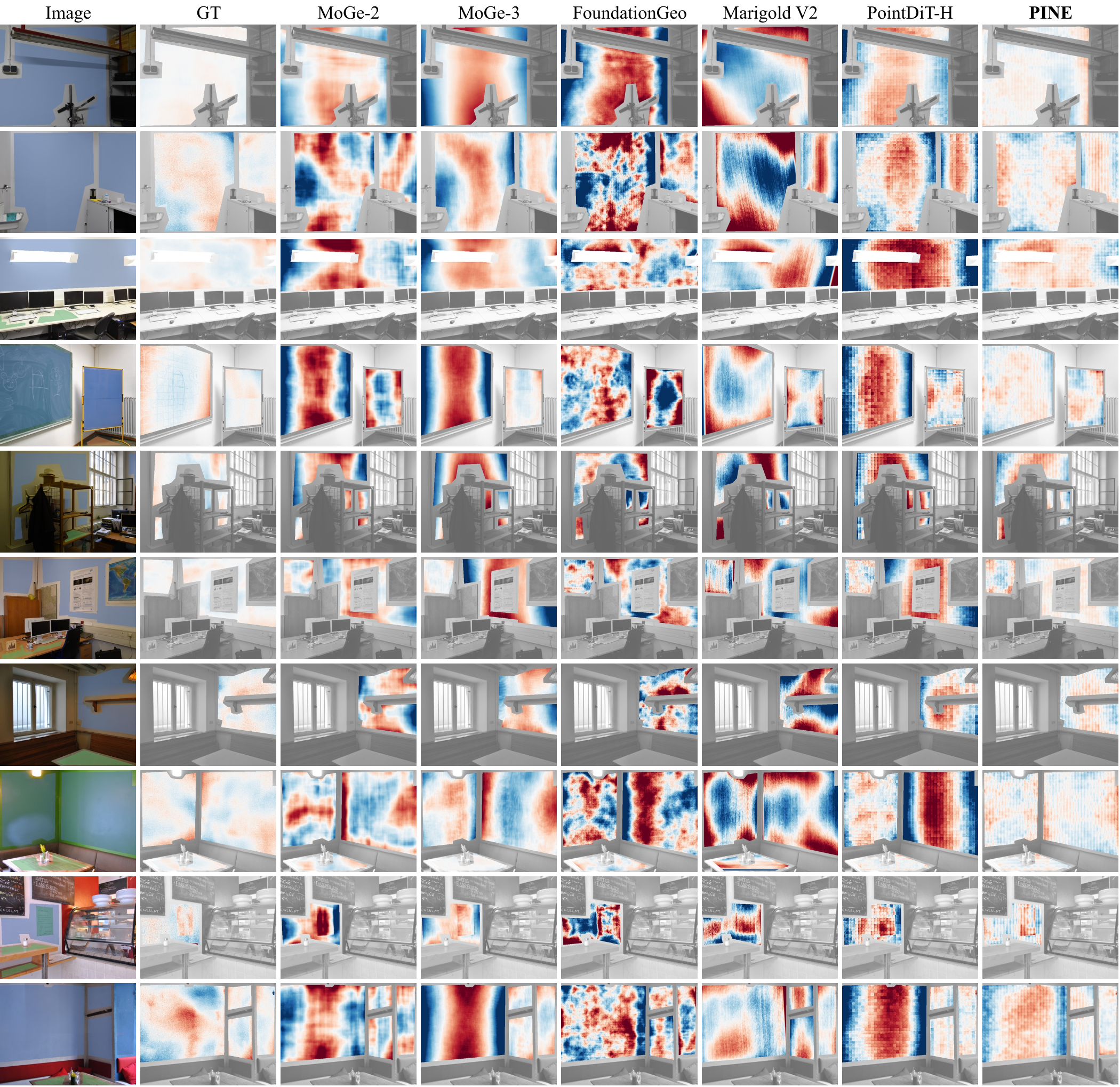}
\caption{\textbf{Planarity on iBims-1.} Each row shows the input and, for the ground truth and each method, the signed distance of each point to the plane fitted to its annotated plane region. Red and blue mark the two sides of the plane, and gray marks pixels outside the planes.}
\label{fig:supp_planarity}
\end{figure*}

\begin{figure*}[tp]
\centering
\includegraphics[width=\textwidth]{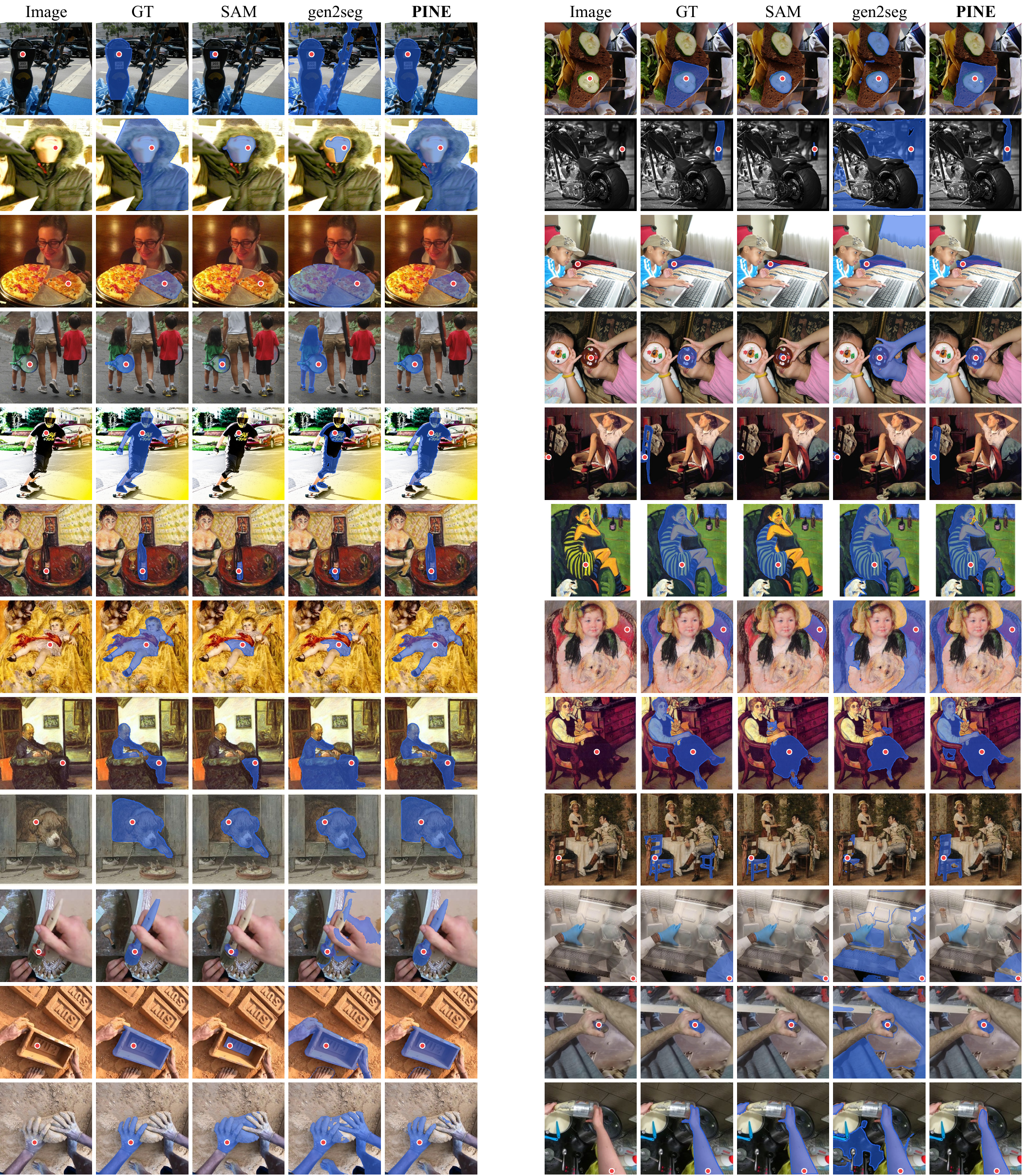}
\caption{\textbf{Point-prompted instances (1/2).} Each row shows the input with its point prompt, the ground-truth instance and the masks of SAM, gen2seg and PINE for the same prompt.}
\label{fig:supp_instance_1}
\end{figure*}

\begin{figure*}[tp]
\centering
\includegraphics[width=\textwidth]{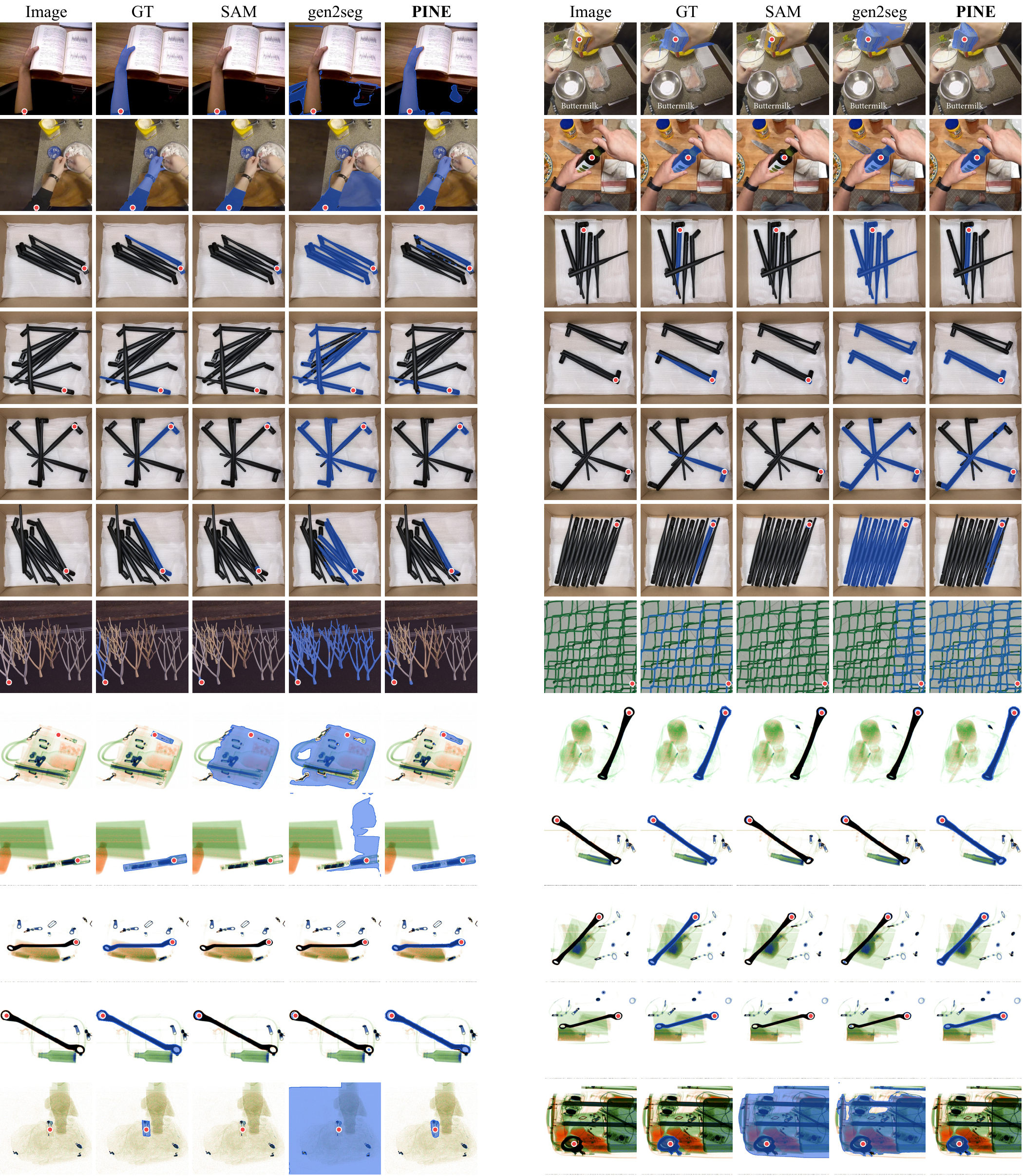}
\caption{\textbf{Point-prompted instances (2/2).} Each row shows the input with its point prompt, the ground-truth instance and the masks of SAM, gen2seg and PINE for the same prompt.}
\label{fig:supp_instance_2}
\end{figure*}

This section complements \cref{fig:qualitative} of the main paper with more examples.
\Cref{fig:supp_point_err_1,fig:supp_point_err_2,fig:supp_point_err_3} compare the point maps, \cref{fig:supp_depth_err} the depth, \cref{fig:supp_normal_err} the surface normals, \cref{fig:supp_planarity} the planes of iBims-1, and \cref{fig:supp_instance_1,fig:supp_instance_2} the point-prompted instances of PINE and the baselines.

\clearpage

